\documentclass[preprint,9pt,authornum]{elsarticle}
\usepackage{graphicx}
\usepackage{amssymb}
\usepackage{amsmath}
\usepackage{longtable}
\usepackage{algorithm} 
\usepackage{algpseudocode} 
\usepackage{comment}
\usepackage{amsmath}
\usepackage{subcaption}
\usepackage{graphicx}
\usepackage{algorithm}
\usepackage{hyperref}
\usepackage{mathtools}
\usepackage{amssymb}
\usepackage{color}
\usepackage{lineno}
\usepackage{graphicx}
\usepackage{algorithm}
\usepackage{amssymb}
\usepackage{amsmath}
\usepackage{makecell}
\usepackage{algorithm} 
\usepackage{algpseudocode} 
\usepackage{pdflscape}
\usepackage{adjustbox}
\usepackage[utf8]{inputenc}
\DeclareUnicodeCharacter{00B2}{\ensuremath{{}^2}}
\usepackage[edges]{forest}
\usetikzlibrary{arrows.meta}
\usetikzlibrary{calc}

\usepackage[figuresright]{rotating}
\usepackage{lineno}

\journal{Applied Soft Computing, Elsevier}

\begin{document}

\begin{frontmatter}
\title{Random vector functional link network: recent developments, applications, and future directions}
\author[label1]{A. K.  Malik}
\ead{phd1801241003@iiti.ac.in}
\address[label1]{Department of Mathematics, Indian Institute of Technology Indore, Simrol, Indore, 453552, India}

\author[labelG]{Ruobin Gao}
\ead{gaor0009@e.ntu.edu.sg}
\address[labelG]{School of Civil \& Environmental Engineering, Nanyang Technological University, Singapore}

\author[labelus]{M.A. Ganaie}
\ead{mudasirg@umich.edu}

\address[labelus]{Department of Robotics, University of Michigan, Ann Arbor, MI 48109, USA\fnref{label44}}

\author[label1]{M. Tanveer\corref{cor1}}
\cortext[cor1]{Corresponding authors}
\ead{mtanveer@iiti.ac.in}

\author[label2,label3]{Ponnuthurai Nagaratnam Suganthan\corref{cor1}}
\ead{p.n.suganthan@qu.edu.qa}
\address[label2]{KINDI Center for Computing Research,
College of Engineering, 
Qatar University,
Doha, Qatar}
\address[label3]{School of Electrical \& Electronic Engineering, Nanyang Technological University, Singapore\fnref{label4}}

\begin{abstract}
Neural networks have been successfully employed in various domains such as classification, regression and clustering, etc. Generally, the back propagation (BP) based iterative approaches are used to train the neural networks, however, it  results in the issues of  local minima, sensitivity to learning rate and slow convergence. To overcome these issues, randomization based neural networks such as random vector functional link (RVFL) network have been proposed. RVFL model has several characteristics such as fast training speed, direct links, simple architecture, and universal approximation capability, that make it a viable randomized neural network. This article presents the first comprehensive review of the evolution of RVFL model, which can serve as the extensive summary for the beginners as well as practitioners. 
We discuss the shallow RVFLs, ensemble RVFLs, deep RVFLs and ensemble deep RVFL models. The variations, improvements and applications of RVFL models are discussed in detail. Moreover, we discuss the different hyperparameter optimization  techniques followed in the literature to improve the generalization performance of the RVFL model. Finally, we give potential future research directions/opportunities that can inspire the researchers to improve the RVFL's architecture and learning algorithm further.    
\end{abstract}

\end{frontmatter}


\section{Introduction}
Artificial intelligence (AI) is a rapidly growing field that has the potential to transform many aspects of the world \citep{zhang2021study}. It refers to the development of computer systems that can perform tasks that typically require human intelligence, such as learning, decision making, and problem solving \citep{russell2017artificial}. AI has the potential to revolutionize industries and improve efficiencies in a wide range of fields, including healthcare \citep{gao2022inpatient}, transportation \citep{li2021spatial}, finance \citep{giudici2021shapley}, and energy \citep{gao2023dynamic}. Machine learning algorithms are the engine of AI. Hence, developing advanced machine learning algorithms for various tasks is of real value.

Among machine learning algorithms, the artificial neural networks (ANNs) have received considerable attention due to their success in diverse domains such as medicine \cite{lisboa2002review}, chemistry \cite{himmelblau2000applications}, robotics \cite{king1989neural}, control systems \cite{hunt1992neural}, industrial applications and function approximation \cite{chen1990non,csaji2001approximation}, so on. The architecture of ANN is inspired by the topology of biological neurons \cite{basheer2000artificial}. The ANN consists of neurons which are simple processing units and these neurons are connected via weighted links. The neurons do mathematical operations that are either linear or nonlinear, and they carry out some task that enables the artificial neural network (ANN) to approximate the unknown function (rule) that generates the data \cite{leshno1993multilayer}. The training phase of an ANN is an iterative process, and all the parameters are tuned via the back propagation (BP) method \cite{chauvin2013backpropagation}. The traditional iterative techniques based on the BP algorithm have some shortcomings, i.e., slow convergence \cite{jacobs1988increased}, not getting global minima that leads to generates sub-optimal parameters \cite{gori1992problem}, and very sensitive to learning rate \cite{magoulas1999improving}.

To overcome the above-aforementioned issues, randomization based techniques have been proposed \cite{zhang2016survey,suganthan2021origins} with fast convergence and universal approximation properties. 
The single hidden layer feed-forward neural (SLFN) network architectures have been extensively studied in the last twenty years and employed in various domains, e.g., classification and regression problem, due to their universal approximation capability \cite{park1991universal,leshno1993multilayer,hornik1989multilayer,scarselli1998universal,zhang1995wavelet}. 
In most of the SLFN networks, the learning process is done in output layer while as weights and biases are generated randomly in hidden layer. The output layer weights are calculated either via closed form solution \cite{pao1994learning} or iterative process \cite{beck2009fast}. The origin of randomized feedforward networks can be traced in late 20th century \cite{suganthan2021origins}.
In 1988, Broomhead and Lowe \cite{broomhead1988radial} discussed universal approximation property using radial basis function (RBF) network with random centers \cite{cao2018review}.
There are several other architectures \cite{huang2022graph} like RBF network, recurrent neural network (RNN) which have randomization based training algorithms \cite{zhang2016survey}.

Moreover, Schmidt et al. \cite{schmidt1992feed} proposed  feed-forward neural network with random weights based on randomization technique. At the same time, connecting the input layer to the output layer via direct link, Pao et al. \cite{pao1992functional,pao1992neural} proposed random vector functional link (RVFL) neural network in 1992, wherein the parameters (weights and biases) from the input layer to the hidden layer are generated randomly from a fixed domain and the output weights are need to be computed analytically. The idea of the direct link can be traced back to the pioneering work in fuzzy systems \citep{takagi1985fuzzy} in 1985. The direct links have shown significant improvement in RBF networks' performance \citep{looney2002radial} in 2002. The intention behind creating a direct link between inputs and outputs is to capture information about how the first derivative of the output with respect to the inputs. In 1994, Igelnik and Pao proved that RVFL is a universal approximator \cite{igelnik1995stochastic}. Recently, Needell et al. \cite{needell2020random} proposed RVFL networks for function approximation on manifolds that fill the theoretical gap lacking in \cite{igelnik1995stochastic}. In 2016, Zhang and Suganthan \cite{zhang2016comprehensive} conducted a comprehensive evaluation of RVFL for classification problems and concluded some remarkable results about this architecture. After that, the RVFL model got the attention of researchers from diverse domains due to its simple architecture, fast training speed, and universal approximation capability. The shallow RVFL model has been employed successfully in several domains, i.e., forecasting \cite{ren2016random,bisoi2019modes,del2021randomization}, non-linear system identification \cite{chakravorti2020non}, function approximation \cite{igelnik1995stochastic}, classification  and regression problem \cite{zhang2019unsupervised,cao2017impact,vukovic2018comprehensive}, etc.

The RVFL model transforms the original feature space to randomized feature space via random feature map (RFM), and this randomization process makes the RVFL model an unstable classifier. Ensemble learning techniques develop stable, robust, and accurate model integrating the several models known as base models \cite{ren2016ensemble}. Combining the diverse and accurate base models \cite{dong2020survey}, the ensemble model performs better than its constitutes models. Broadly speaking, the ensemble learning can be divided into three categories, i.e., bagging \cite{breiman1996bagging}, boosting \cite{freund1996experiments} and stacking framework \cite{wolpert1992stacked}. Thus, the RVFL model has been improved (developed) in ensemble frameworks, and the more stable and robust RVFL variants have been proposed and employed in various domains, i.e., crude oil price forecasting \cite{zhang2020novel}, medical domain \cite{shi2018cascaded}, and classification problem \cite{katuwal2018enhancing}, etc. Deep learning architectures have the high representation learning capability due to  several stacked layers for extracting informative features \cite{gallicchio2020deep} and have been successfully employed in several domains, i.e., computer vision \cite{voulodimos2018deep}, bioinformatics \cite{li2019deep}, and visual tracking  \cite{marvasti2021deep}, and speech recognition task \cite{nassif2019speech} etc. On the other hand, Utilising the strength of two individual growing fields, i.e., ensemble learning and deep learning, researchers are developing ensemble deep models \cite{cao2020ensemble,ganaie2021ensemble}. The shallow RVFL model has been extended to deep and ensemble deep architectures that improve its generalization performance. The deep RVFL network \cite{shi2021random} has several stacked layers wherein all parameters of hidden layers are generated randomly and kept fixed during the training process, and only output layer parameters are needed to be computed analytically. The deep RVFL model has better representation learning compared to the shallow RVFL model. The deep RVFL model faces memory issues when training data size, the no. of hidden layers, and the feature dimension of the data are considerable. Therefore, to address these issues, an implicit ensemble technique based ensemble deep RVFL network known as edRVFL model has been proposed  \cite{shi2021random}.

RVFL has been improved in multiple aspects both in shallow and deep frameworks and has been applied in diverse domains. In this paper, we present journey of  shallow and deep RVFL along with its applications. We conclude this article with potential future research directions that might inspire researchers to develop this architecture further. 

The rest part of this paper is organised as follows. In Section \ref{Sec: it's mathematical foundation}, we present the formulation of the standard RVFL model. Section \ref{Sec:Research methodology} discusses the research methodology and objective of this article. The improvements in shallow RVFL and their applications are discussed in the Section \ref{Sec:Developments and applications of RVFL model}.  In Section \ref{Sec:Semi-supervised methods based on RVFL model} and \ref{Sec:Clustering methods based on RVFL model}, we discuss the  semi-supervised methods  and clustering based methods that have been developed based on the RVFL model, respectively. We present the ensemble learning based RVFL model in Section \ref{Sec:Ensemble frameworks based on RVFL model} and Section \ref{Sec:Deep architectures based on RVFL model} discusses the deep architectures based on the RVFL model. Section \ref{Sec:Hyper-parameters optimization and experimental setup} discusses the hyper-parameters optimization and experimental setup details and Section \ref{Sec:Other applications of RVFL model} discusses the applications of the RVFL model. In Section \ref{Sec:comparing with SORT}, comparison of RVFL with other machine learning models are given. Finally, the potential future directions with conclusions are given in Section \ref{Sec:Conclusion}.

\section{The standard RVFL architecture and it's mathematical foundation}
\label{Sec: it's mathematical foundation}
In this section, we discuss formulation of the standard RVFL model.
Let $X=[x_{1},x_{2},\cdots,x_{N}]^{T},~~ x_{i} \in \mathbb{R}^{d}$ be the training dataset and $Y=[y_{1},y_{2},\cdots,y_{N}]^{T},~~ y_{i} \in \mathbb{R}^{c}$, be the target matrix. Here, $d$ represents the number of features in each sample $(x_{i})$ and $c$ denotes the number of classes. Fig. \ref{fig:Layout of paper} shows the layout of the paper and Fig. 1(b) shows the different types of architectures of the RVFL model. 
\subsection{Random vector functional link (RVFL) network }
\begin{figure*}
\centering
	\subcaptionbox{Shallow RVFL: Red lines show the direct links between input layer to output layer, black lines show the connection between input layer to hidden layer and blue lines show the links between hidden layer to output layer.}{
\includegraphics[width=.6\textwidth]{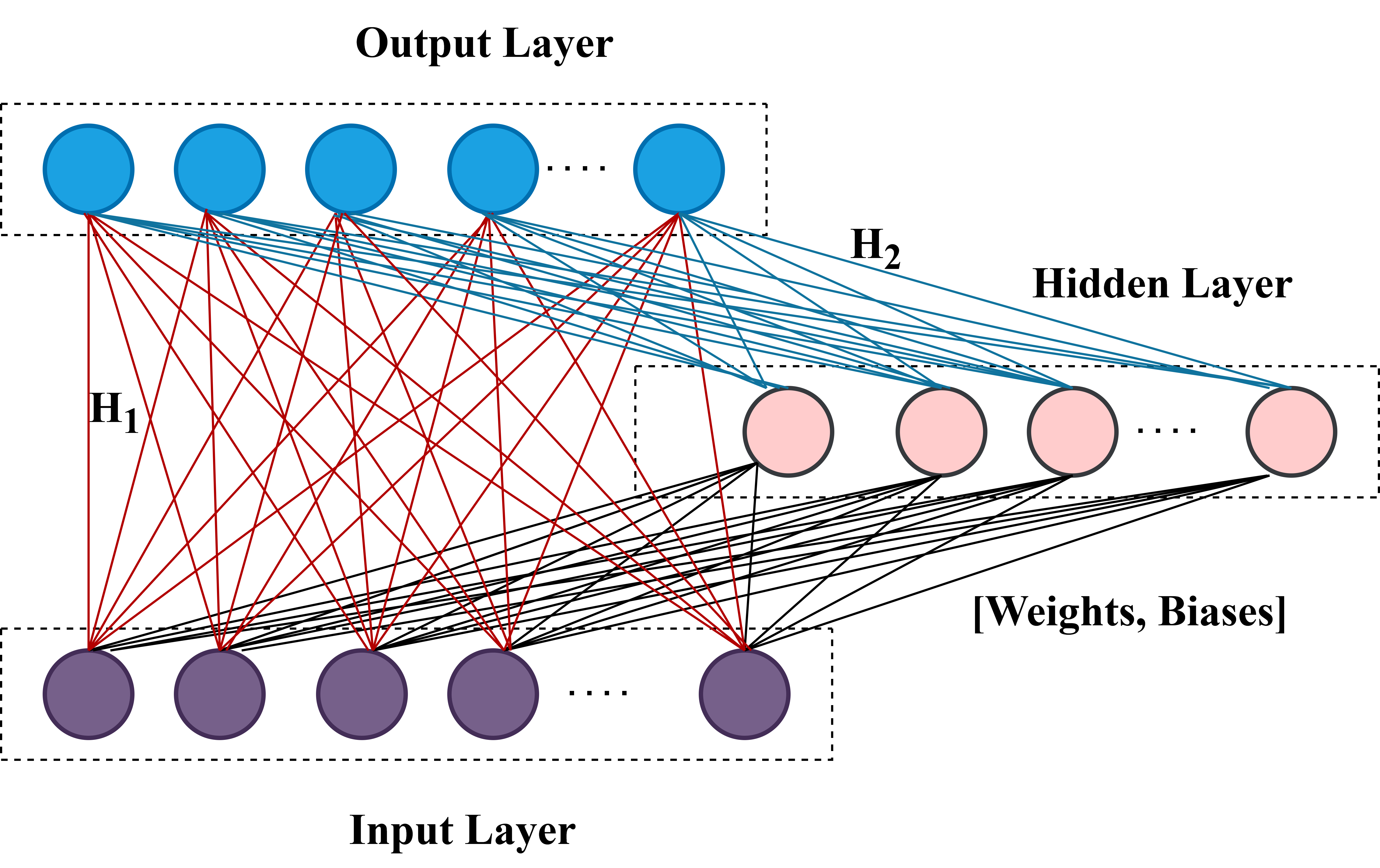}}
\vfill
\vspace{2mm}
\subcaptionbox{Different types of architectures of RVFL model.}{
	\includegraphics[width=.6\textwidth]{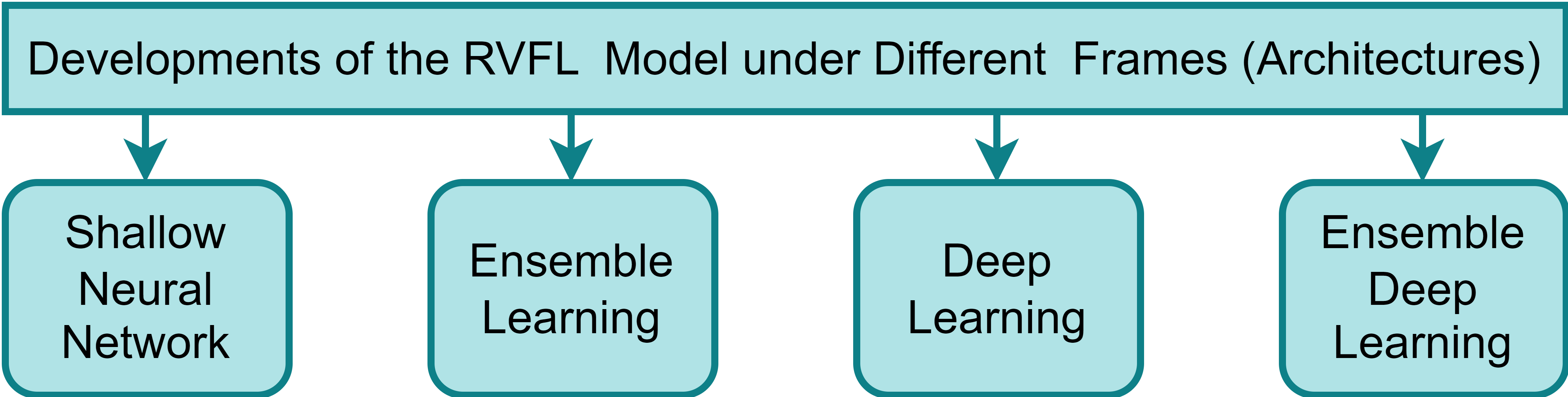}}
		\caption{The architectures of RVFL model.}
		\label{ref:RVFL architecture}
\end{figure*}

 RVFL is a randomized version of a single hidden layer feedforward neural (SLFN) network, with three layers known as the input, hidden, and output layers. All three layers consist of neurons are connected via weights. To avoid the implementation of the back propagation algorithm, the weights from the input layer to the hidden layer are generated randomly from a domain and kept fixed during the training process. Only the output weights are analytically computed by the least square method. In RVFL, original features are also used to link the input and output layers. The direct links improve the generalization performance of RVFL \cite{zhang2016comprehensive}. The architecture of RVFL is given in Fig.\ref{ref:RVFL architecture} (a). Mathematically, the RVFL model, i.e., $\mathbf{f}:\mathbb{R}^{d} \rightarrow \mathbb{R}^{c}$, can be written as:
 \begin{equation}
    \mathbf{f}(x_{i})=\sum_{k=1}^{d}\beta_{k}{x}_{ik}+\sum_{k=d+1}^{L}\beta_{k}\theta(\langle \mu_{k},x_{i}\rangle+\sigma_{k}),~i=1,2,\cdots,N. 
 \end{equation}
 
In particular, $\langle \mu_{k},x_{i}\rangle=\mu_{k}\cdot x_{i}$ is the standard inner product defined on Euclidean space ($\mathbb{R}^{d}$).
The objective function of standard RVFL model  with $L$ hidden nodes can be written as:

\begin{align}
\label{eqn:RVFL1}
& \text{min}~~\frac{1}{2}\Vert\beta\Vert^{2}+ \frac{1}{2}C\Vert  \xi\Vert^{2}\nonumber\\ & \text{subject to}~ H\beta- Y=\xi.
\end{align}


where $\left\|.\right\|$ represents the Frobenius norm and $\xi=[\xi_{1},\xi_{2},\ldots,\xi_{N}]^{T}$ is the error term corresponding to $N$ samples. This is the quadratic optimization problem with linear constraints. $\beta$ and $H$ are the output weight matrix and concatenation matrix consist of input data and outputs from the hidden layer, respectively and $Y$ is the target matrix.

The optimization problem \eqref{eqn:RVFL1} can be rewritten as: 
\begin{align}
\label{eqn:RVFL11}
\min_{\beta \in \mathbb{R}^{(d+L)\times c}} \frac{1}{2}\left\|\beta\right\|^{2}+\frac{1}{2}C\left \|H\beta-Y\right\|^{2},
\end{align}

here,
$$H=[H_1~ H_2]_{N \times (d+L)},$$
where
\begin{align}
 H_1= \begin{bmatrix} 
x_{11}&x_{12}&\cdots & x_{1d} \\
\vdots&\vdots & \ddots & \vdots\\
x_{N1} &x_{N2}&\cdots & x_{Nd}\\
\end{bmatrix}_{N\times d},
\end{align}
and
\begin{align}
 H_2= \begin{bmatrix} 
\theta(\mu_{1}\cdot x_{1}+\sigma_{1})&\theta(\mu_{2}\cdot x_{1}+\sigma_{2})&\dots &\theta(\mu_{L}\cdot x_{1}+\sigma_{L})   \\
\vdots & \ddots & \vdots&\vdots\\
\theta(\mu_{1}\cdot x_{N}+\sigma_{1}) & \theta(\mu_{2}\cdot x_{N}+\sigma_{2})&\dots & \theta(\mu_{L}\cdot x_{N}+\sigma_{L})\\
\end{bmatrix}_{N\times L},
\end{align}
 $$\beta=\begin{bmatrix}
\beta_1\\
\beta_2\\
\vdots\\
\beta_{(d+L)}
\end{bmatrix}_{(d+L)\times c}
\text{and}~~
Y=\begin{bmatrix}
y_{1}\\
y_{2}\\
\vdots\\
y_{N}
\end{bmatrix}_{N\times c}.$$\\\\
Here, $\beta_{k}=[\beta_{k1},\beta_{k2},\cdots,\beta_{kc}]$ is the output weight vector connecting the $k^{th}$  input (hidden) node and the output nodes, where $1 \leq k \leq d+L$, and ${\mu}_{j}=[\mu_{j1},\mu_{j2},\cdots,\mu_{jd}]$ is the weight vector connecting the $j^{th}$ hidden node and the input nodes, $1 \leq j \leq L$ . Also, $x_{i}=[x_{i1},x_{i2},\cdots,x_{id}]$ is the $i^{th}$ sample. For the target matrix, $y_{i}=[y_{i1},y_{i2},\cdots,y_{ic}],$ $1 \leq i \leq N$. Moreover, $\theta(\cdot)$ and $\sigma_{i}$ are the non-constant activation function and the bias term of $i^{th}$ hidden node, respectively.

The optimal solution of the problem \eqref{eqn:RVFL1} when $\delta=\frac{1}{C} =0$ is as follows:
\begin{align}
\beta=H^{+}Y,
\end{align}

where $H^{+}$ represents the Moore-penrose generalized inverse of the matrix $H$  \cite{rao1971further}. The regularization term is employed to avoid the over-fitting issue.

Therefore, the optimization problem with regularization term is solved. Let the Lagrangian be

\begin{align}
L(\beta,\xi,\alpha)=C\frac{1}{2}\left\|\xi\right\|^{2} + \frac{1}{2}\left\|\beta\right\|^{2}-\alpha^{T}(H\beta-Y-\xi),
\end{align}
 and obtain the partial derivatives of L w.r.t $\beta$, $\xi$ and $\alpha$ and set them to zero.
\begin{align}
\label{eq:beta}
 \frac{\partial L}{\partial \beta}=0 \implies \beta=H^{T}\alpha,
\end{align}
\begin{align}
\label{eq:alpha}
 \frac{\partial L}{\partial \xi}=0 \implies \alpha=-C\xi,
\end{align}
\begin{align}
\label{eq:beta_and_H}
 \frac{\partial L}{\partial \alpha}=0 \implies H\beta-Y-\xi=0.
\end{align}
When the number of features is less than number of samples, from \eqref{eq:alpha} and  \eqref{eq:beta_and_H},  we obtain $\alpha=-C(H\beta-Y)$. By substituting the value of $\alpha$ in ~\eqref{eq:beta}:
\begin{align}
    \beta= (H^{T}H+\frac{1}{C}I)^{-1}H^{T}Y
\end{align}
When the number of samples is less than the number of features, after substituting  \eqref{eq:beta} and \eqref{eq:alpha} into \eqref{eq:beta_and_H}, we obtain $\alpha=(HH^{T}+\frac{1}{C}I)^{-1}Y$, By substituting the value of $\alpha$ in \eqref{eq:beta}:
\begin{align}
    \beta=  H^{T}(HH^{T}+\frac{1}{C}I)^{-1}Y.
\end{align}
Therefore, in this case the optimal solution of \eqref{eqn:RVFL1} is  given by,
\begin{align}
\label{eq:finalsolution}
\beta = \left\{
        \begin{array}{ll}
            (H^{T}H+\frac{1}{C}I)^{-1}H^{T}Y, ~ \quad (d+L)\leq N\\
           H^{T}(HH^{T}+\frac{1}{C}I)^{-1}Y, ~\quad  N < (d+L),
        \end{array}
    \right.
\end{align}
where $C$ is the regularization parameter to be tuned and $I$ is an identity matrix of appropriate dimension. Both matrices $H^{T}H$ and $HH^{T}$ are symmetric positive semidefinite matrix and $C > 0$, so both matrices in (\ref{eq:finalsolution}) is positive definite, therefore, $(HH^{T}+\frac{1}{C}I)$ and $(H^{T}H+\frac{1}{C}I)$ are non singular matrix.

\begin{figure*}
\centering
	{
\includegraphics[width=.8\textwidth]{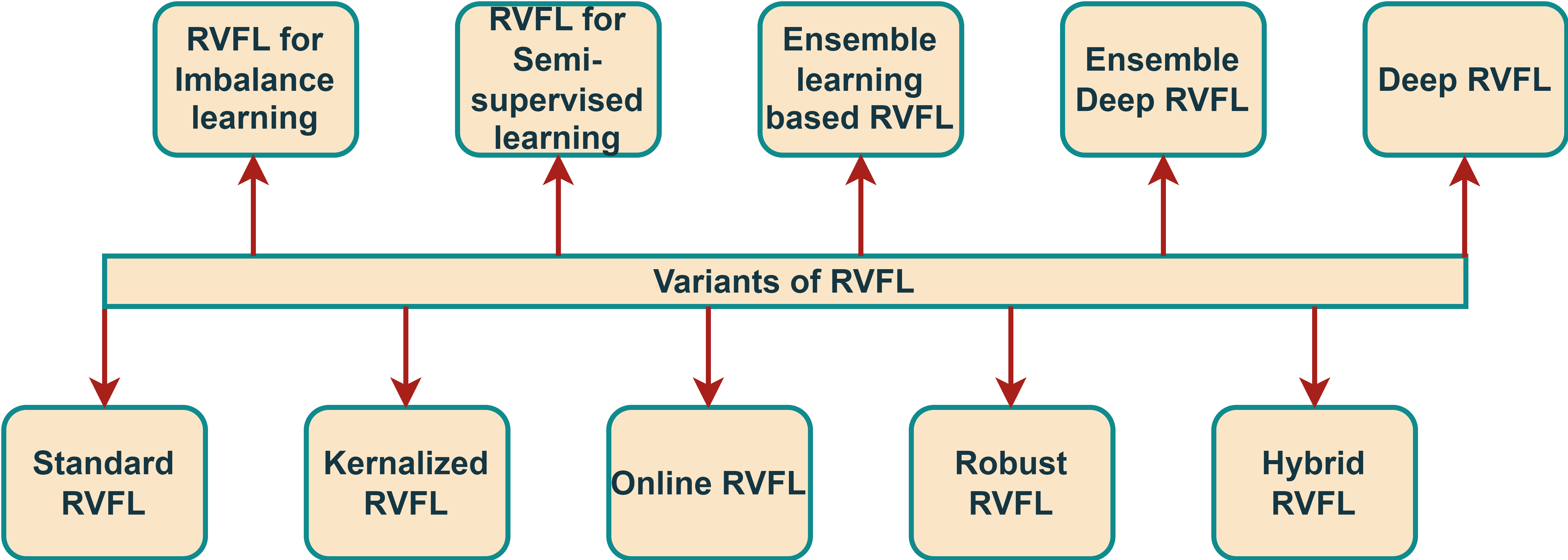}}
		\caption{Architectural/Algorithmic Variants of the RVFL model.}
		\label{ref:RVFL_Variants}
\end{figure*}

\begin{landscape}
 \begin{figure}
	\centering
	\begin{adjustbox}{max size={\paperheight}{1\textheight}}
\begin{forest}
	forked edges,
	/tikz/>/.tip={Stealth[]},
	my label/.style={%
		tikz+={\node [font=\scriptsize, anchor=south east] at (.north east) {\#1};}
	},
	for tree={%
		draw,
		align=center,
		minimum height=20mm,
		minimum width=30mm,
		anchor=center,
		l sep'=15mm,
		edge={->},
		s sep'=5mm,
		    if n children=0{
		    	no edge, , rounded corners,	draw=white, align=center, minimum height=2mm
		}{},
	}
	[\Large{\textbf{
	Introduction and theory of RVFL [1-2], Research methodology [3]}},
	[\Large{\textbf{Architectural and algorithmic variants}}, l sep'=20mm
	[\Large{\textbf{Shallow RVFL}},thick, l sep'=10mm
	[\parbox{15em}{
		    	\begin{itemize}
		    		\item \Large{\textbf{Empirical evaluation of RVFL [4.1]}}
		    		\item \Large{\textbf{Weight initialization techniques based RVFL [4.2]}}
		    		\item \Large{ \textbf{RVFL with manifold learning [4.3]}}
		    		\item \Large{\textbf{Robust RVFL [4.4]}}
		    		\item \Large{\textbf{Kernelized RVFL [4.5]}}
		    		\item \Large{\textbf{Other techniques based RVFL [4.6]}}
		    		\item \Large{\textbf{Imbalance learning based on RVFL [4.7]}}
		    		\item \Large{\textbf{Multi-label classification based on RVFL [4.8]}}
		    	\end{itemize}},edge+={densely dashed},thick,l sep'=8mm	
	    						[
	    						 ]
	    						 ]
	    						 ]
    [\Large{\textbf{RVFL with ensemble learning}},thick, l sep'=10mm
	[\parbox{15em}{
	               \begin{itemize}
		        	\item \Large{\textbf{Bagging technique based RVFL [7.1]}}
		        	\item \Large{\textbf{Boosting technique based RVFL [7.2]}}
		        	\item \Large{\textbf{Stacking technique based RVFL [7.3]}}
			    	\item \Large{\textbf{Ensemble RVFL-based on decomposition [7.4]}}
		        	\end{itemize}},edge+={densely dashed},thick,l sep'=8mm
				[
				 ]
				 ]
				 ]
	[\Large{\textbf{Deep RVFL architectures}},thick, l sep'=10mm
	[\parbox{15em}{
                   \begin{itemize}
			      \item \Large{\textbf{Stacked deep RVFL [8.1]}}
			      \item \Large{\textbf{Ensemble deep RVFL [8.2]}}
		          \item \Large{\textbf{Hybrid deep RVFL [8.3]}}
				  \end{itemize}},edge+={densely dashed},thick,l sep'=8mm
				[
				 ]
				 ]
				 ]
    [\Large{\textbf{Un-/semi-supervised}},thick, l sep'=10mm
	[\parbox{15em}{
                  \begin{itemize}
                   \item \Large{\textbf{Semi-supervised based on RVFL [5]}}
			      \item \Large{\textbf{Clustering methods based on RVFL [6]}}
			 	\end{itemize}},edge+={densely dashed},thick,l sep'=8mm
				[
				 ]
			     ]
			     ]   
				 ]
	[\Large{\textbf{Hyper-parameter optimization}}\\\Large{\textbf{and experimental setup}},thick,l sep'=5mm
	[\parbox{12em}{
                   \begin{itemize}
				  \item \Large{\textbf{Hyper-parameter optimization for single layer RVFL [9.1]}}
					\item \Large{\textbf{Hyper-parameter optimization for deep RVFL [9.2]}}
					\item \Large{\textbf{Experimental setup [9.3]}}
				\end{itemize}},edge+={densely dashed},thick,l sep'=7mm
				[
				 ]
				 ]
				 ]
	[\Large{\textbf{Applications of RVFL}},thick,l sep'=10mm
	[\parbox{12em}{
                  \begin{itemize}
				 \item \Large{\textbf{Electricity load forecasting [10.1]}}
				  \item \Large{\textbf{Solar power forecasting [10.2]}}
		     	\item \Large{\textbf{Wind power forecasting [10.3]}}
		    	  \item \Large{\textbf{Financial time series forecasting [10.4]}}
				\end{itemize}},edge+={densely dashed},thick,l sep'=8mm
				[
				 ]
				 ]
				 ]
	[\Large{\textbf{Comparison with other }} \\ \Large{\textbf{ state of the art}}\\ \Large{\textbf{ ML techniques [11]}}\\ \Large{\textbf{Conclusions and future directions [12]}}, l sep'=13mm
	[\parbox{12em}{}
	 ]
	 ]
	 ]
  ] ]
	\end{forest}
\end{adjustbox}
\caption{Layout of the paper.} \label{fig:Layout of paper}
\end{figure}
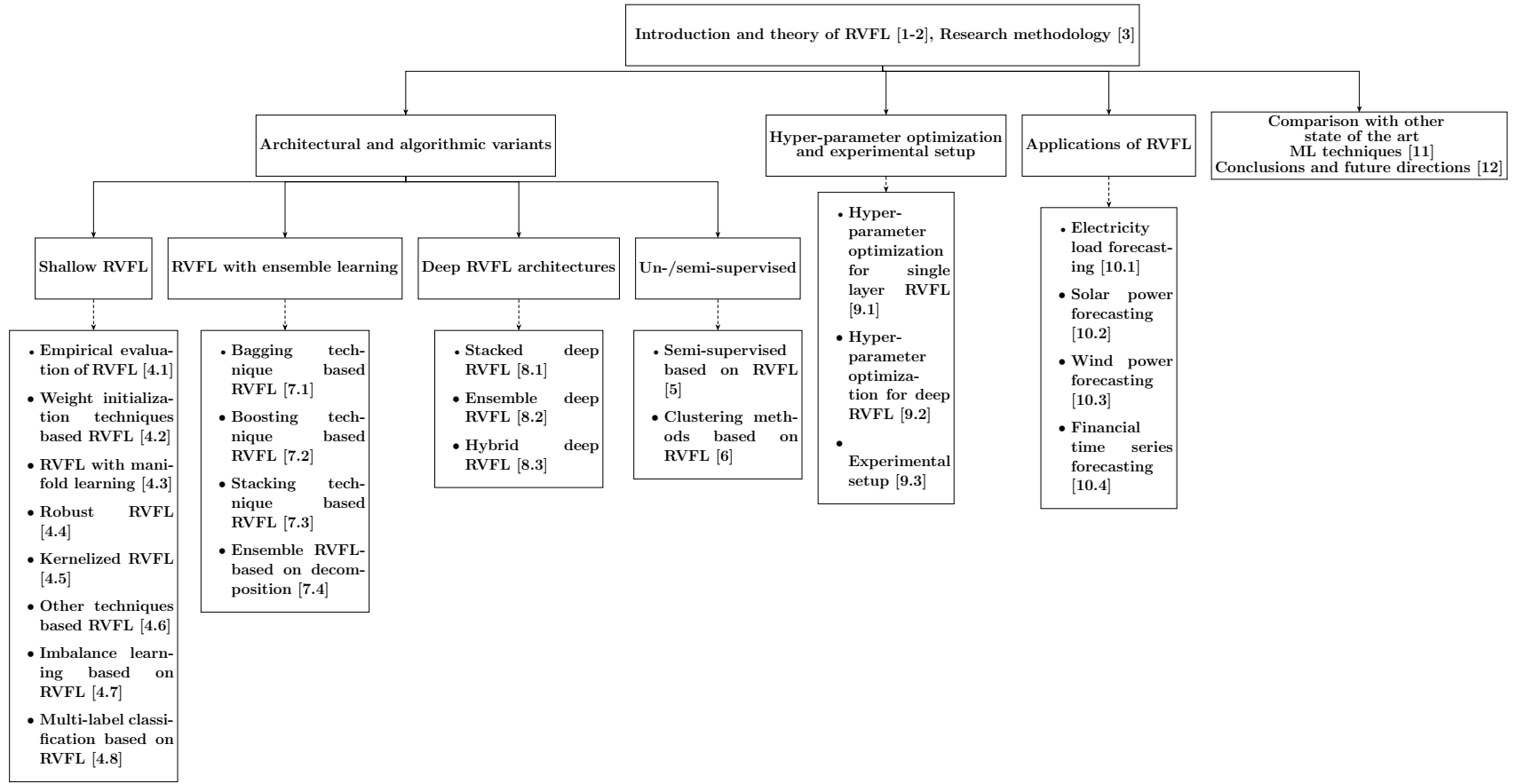
\end{landscape}

\section{Research methodology}
\label{Sec:Research methodology}
The studies included in this review paper were obtained by searching the Google Scholar and Scopus search engines.
The papers are the result of keywords random vector functional link and  deep RVFL. The articles are screened based on the title and abstract, followed by the screening of full-text version. The focus of this article is to represent the developments of shallow RVFL, ensembles of RVFL models, deep RVFL, ensembles of deep RVFL-based models and their applications. Having these search strategy, we included fundamental papers on RVFL
model proposed in the late 20th century and after that we included all
RVFL papers from 2016 onward. This paper explains the following issues:
\begin{enumerate}
    \item How the weights are initialized in RVFL-based models.
    \item Techniques used to improve the robustness of the models in presence of noise/outliers and class imbalance problems.
    \item Techniques followed for selection of hidden nodes like kernelized approach for RVFL.
    \item How different approaches of ensemble learning like bagging, boosting and stacking improve the performance of the RVFL-based ensemble models.
    \item Techniques followed for the development of deep RVFL-based models.
    \item Development of RVFL-based models for different scenarios like semi-supervised learning, unsupervised learning and regression problems.
    \item How to select the different parameters of the RVFL-based models, followed by the techniques for the optimization of these models.
    \item Approaches followed for analyzing the effect of hyperparameters on the generalization performance of the RVFL-based models.
    \item The loop holes in the current literature of the RVFL-based models and possible future research directions.
\end{enumerate}

\section{Developments and applications of RVFL model}
\label{Sec:Developments and applications of RVFL model}
In this section, we discuss the journey of improvements in the shallow RVFL architecture.
\subsection{Empirical evaluation of RVFL for classification and regression problems}
\label{Sec:Empirical evaluation of RVFL for classification and regression problems}
Few scientific questions regarding the RVFL model have been addressed rigorously in the literature. Zhang and Suganthan \cite{zhang2016comprehensive} conducted a comprehensive evaluation of RVFL model to answer the following questions, i.e., the impact of direct links, the effect of bias term in the output layer, and different types of activation functions in the hidden layer, domain from where random parameters are generated and the output weights calculation techniques over 121 UCI classification datasets. The experimental results demonstrate that RVFL with direct links from the input layer to the output layer has better generalization performance than RVFL without direct links, and output weights calculated via ridge regression method are better than Moore-Penrose pseudoinverse based output weights. Moreover, generating the random weights and biases within a suitable domain from the input layer to the hidden layer has a significant impact on the performance of RVFL, and biases term in the output layer may or may not have any impact on the performance of RVFL. Similarly, a comprehensive evaluation of orthogonal polynomial expanded RVFL (OPE-RVFL) \cite{vukovic2018comprehensive} method was conducted in which the input patterns are expanded non-linearly using four orthogonal polynomials, i.e., Chebyshev, Hermite, Laguerre, and Legendre with three activation functions tansig, logsig, and tribas. The results demonstrate that direct links in OPE-RVFL model also have a significant impact on regression problem, and ridge regression based approach is better than Moore-Penrose pseudoinverse based approach. Basically, this study also supports direct links in RVFL (as in \cite{zhang2016comprehensive}). Moreover, the results demonstrate that the OPE-RVFL model with Chebyshev polynomial performs better than other compared polynomials, and tansig is the best performing activation function followed by tribas and logsig. Table \ref{Table:activation function} shows the commonly used activation functions with RVFL model in the literature.

The random feature mapping (RFM) mechanism has a significant impact on the performance of RVFL and plays a vital role in the success of RVFL, but there is very little research to explore this topic. Cao et al. \cite{cao2020study} conducted an experiment to study the relationship between the rank of input data and the performance of RVFL and introduced a concept to measure the quality of RFM via dispersion degree of matrix information distribution (DDMID). Moreover, in \cite{cao2020study}, several scientific questions were addressed, such as the relationship between the performance of the RVFL model and the rank of the input dataset, the sensitivity of this relationship with the different types of activation function, and the number of hidden nodes, respectively and many more. Rasheed et al. \cite{rasheed2020respiratory} implemented standard RVFL with different activation functions into the respiratory motion prediction and compared RVFL with direct and without direct links. The authors find out that the results with hardlim activation function are better than results with sigmoid, sine, tribas, radbas and sign functions, and direct links prevent RVFL from overfitting. Also, RVFL with direct links has better generalization performance than RVFL without direct links. Dudek \cite{dudek2020direct} also conducted an experiment to answer a few scientific questions related to the RVFL model, such as whether direct links and bias terms in the output layer are necessary for modeling the data in regression problem and what should be the optimal domain (or optimal process) for choosing the random parameters for the hidden layer. In this study, three methods for selecting the random parameters are considered, and the author concludes that direct links seem helpful in modeling the target function with linear regions. The target function with non-linear nature can be modeled with RVFL having no direct connections and bias terms.
\subsection{RVFL with different weight initialization techniques}
\label{Sec:Improvements in RVFL model based on different weight initialization techniques}
The RVFL model has a random initialization process from the input layer to the hidden layer, wherein original feature space is transformed into randomized feature space, and hence, RVFL has breakneck training speed and less tunable parameters. It is also capable to model complex (linear/non-linear) data and has universal approximation property \cite{igelnik1995stochastic}. On the other hand, the randomization mechanism makes RVFL an unstable model. The quality of random weights and hidden biases significantly affect the RVFL model's performance. Traditionally, the random weights and hidden biases are randomly generated with some probability distribution from a fixed domain, i.e. [-1, 1] and [0, 1], respectively. Study \cite{zhang2016comprehensive} reveal that the aforementioned range is not the optimal domain to choose the random parameters. Zhang et al. \cite{zhang2019unsupervised} proposed a sparse pre-trained random vector functional link (SP-RVFL) network in which sparse autoencoder with $L_1$ norm is employed to generate weights and hidden biases from input layer to hidden layer. Sparse auto-encoder
learned superior network parameters than the randomization process. The experimental evaluation on different datasets demonstrate that SP-RVFL outperforms standard RVFL. SP-RVFL model generates sparse weights between input layer to hidden layer that is better as compared to randomly generated weights.

Cao et al. \cite{cao2019initial} studied the relationship between the probability distribution of the features (variables) in the datasets and probability distribution of input weights and hidden biases in a neural network with random weights (NNRW) (in particular with RVFL and ELM). This study generated seven regression datasets with known distributions (Gaussian, Gamma, and Uniform). The input weights and hidden biases are initialized with different distributions, and the models, i.e., RVFL and ELM, are trained, respectively. The experimental results conclude several phenomena, such as the model having input weights and hidden biases from Gaussian distribution can have a faster convergence rate than ones with the Gamma and Uniform distribution. Suppose one or more features follow Gamma distribution, and input weights and hidden biases follow the same distribution. In that case, the corresponding model has a slow convergence rate and faces an over-fitting issue, etc. In another similar study \cite{cao2017impact}, experiments were conducted to study the impact of Gaussian, Uniform, and Gamma distributions on the performance of the RVFL model, and the results suggest that input weights and hidden biases' probability distribution has a significant impact on the performance of RVFL model. This study also confirms that RVFL with direct links performs better than RVFL without direct links.

Tanveer et al. \cite{tanveer2021ensemble} proposed several ways to generate input weights and hidden biases in RVFL, wherein twin bounded support vector machine (SVM), least-square twin SVM, twin k-class SVM, least-square twin k-class SVM, and robust energy-based least square twin SVM models were employed to initialize input weights and hidden biases. The experiments illustrate that the twin bounded SVM-based approach has better generalization performance with a lower model rank (Friedman rank) than other proposed approaches.

In \cite{fan2019image}, the random subspace Fisher linear discriminant (FLD) method generates the random weights and hidden biases wherein important features are assigned higher weights than weights assigned to more minor essential features. Therefore, this approach improves the performance of RVFL. Pan et al. \cite{pan2020random} proposed a novel method, Jaya-RVFL, where a new emerging intuitive optimization technique- Jaya algorithm \cite{rao2016jaya} is employed to optimize the randomization range of input weights and employed to transient stability assessment in power system.
Lu et al. \cite{lu2020classification} proposed three classification methods, i.e., MobileNet-RVFL-CBA, MobileNet-ELM-CBA, and MobileNet-SNN-CBA, respectively, to classify the brain MRI image. In each scenario, first, the MobileNetV2 \cite{Sandler_2018_CVPR} trained over ImageNet data is employed to extract the features from brain MRI images and then to improve the generalization performance of these randomized neural networks (RVFL, ELM, and SNN \cite{schmidt1992feed}), chaotic bat algorithm (CBA) is utilized to optimize the random weights and biases. The empirical evaluation of these models for classifying brain MRI data demonstrates that MobileNet-RVFL-CBA performs better than MobileNet-ELM-CBA and MobileNet-SNN-CBA in terms of sensitivity and overall accuracy. 
In the literature, it can be seen that several approaches such as different distributions \cite{cao2019initial}, auto encoder \cite{zhang2019unsupervised} and SVM models \cite{tanveer2021ensemble} have been used to calculate the weights from input layer to hidden layer. There is no such method which is always adaptable. Therefore, a further research is required to develop the efficient techniques to initialize the weights from input layer to the hidden layer in RVFL.

\subsection{RVFL with manifold learning theory }
\label{Sec:Manifold learning strategies based RVFL}
Maintaining the data's global and local geometric structure while data is proceed via several randomization processes is a challenging task. Manifold learning theory helps to overcome this issue. The standard RVFL model does not consider the geometrical aspect of the data and hence, lose information that leads to lower generalization performance. Li et al. \cite{li2021discriminative} proposed discriminative manifold RVFL neural network (DM-RVFLNN), wherein a soft label matrix is used to enlarge the margin between inter-class samples that makes the DM-RVFLNN model more discriminative, and manifold learning based with-in class similarity graph is used to enhance the compactness and the similarity with-in class samples. Experiments on the rolling bearing fault diagnosis process demonstrate that the DM-RVFLNN model is superior and effective compared to standard RVFL. Considering the topological relationship of samples and to improve the robustness of the RVFL model, sparse Laplacian regularized RVFL neural network with $L_{2,1}$-norm (SLapRVFL) \cite{guo2021assessing} was proposed to assess the dry weight of hemodialysis patients and the experiments demonstrate that SLapRVFL is more robust than standard RVFL. Here, $L_{2,1}$-norm was used with regularization term to obtain sparse output parameters.
The standard RVFL model doesn't consider the within-class/total variance of training data while obtaining the final output parameters; therefore, Ganaie et al. \cite{ganaie2020minimum} proposed two variants of the RVFL model known as total variance minimization based RVFL (Total-Var-RVFL) that employs the complete variance information of the training data in the objective function of standard RVFL, and intraclass variance minimization based RVFL (Class-Var-RVFL) wherein the variance of each sample from its respective class is considered, and hence, both models show better generalization performance than standard RVFL model.

Parija et al. \cite{parija2021deep} proposed minimum variance-based kernel RVFL (MVKRVFL) to identify the seizure and non-seizure epileptic EEG signal. In MVKRVFL, both total variance of the training data and within class variance are minimized to improve the generalization performance of RVFL model. Kernel trick is also employed to avoid the hidden layer nodes and activation function (as these parameters are chosen in RVFL). In MVKRVFL, deep long short-term memory (DLSTM) \cite{hochreiter1997long} is employed to extract the features from epileptic EEG signals. The empirical evaluation of the DLSTM-MVKRVFL model over EEG data demonstrates that it efficiently classifies seizure and non-seizure movement. In \cite {ganaie2021co}, co-trained RVFL (coRVFL) model is proposed in which two feature spaces i.e., randomly projected features and sparse-$l_{1}$ norm autoencoder based features are employed. The coRVFL model utilizes the strength of different features spaces and impove the generalization performance of RVFL model. 
Alzheimer's disease (AD) typically affects the brain's cognitive functions, damages the cells and memory in the brain. Having heterogeneous medical data, it is a challenging task to diagnose it at an early stage. Dai et al. \cite{dai2017healthy} proposed a hybrid model combining features extracted from different modalities and introduced the manifold concept in the RVFL model to enhance the diagnosis process of AD. Adopting the manifold theory in its optimization process, RVFL model has the capability in maintaining the geometrical properties of the data while calculating the final output parameters. Literature indicates that maintaining the local and global geometrical (or statistical) properties of the data, manifold leaning based RVFL models have better generalization performance than standard RVFL. 
  
\subsection{Robust RVFL models}
\label{Sec:Robust RVFL models}
  The standard RVFL model employs $L_{2}$ norm-based loss function that is sensitive to outliers and hence, affects the generalization performance of the model. The standard RVFL doesn't perform well over noisy datasets, and therefore, one needs to handle the noisy datasets with extra attention. Managing noisy data is a challenging task. Thus, several approaches have been introduced to address such problems.
  To deal with datasets having noise or outliers and to reduce the complexity of the model so that the generalization performance of the model can be improved, Cui et al. \cite{cui2017received} proposed RVFL-based approach, wherein a novel feature selection method is introduced to make the RVFL model more efficient and robust based on the augmented Lagrangian method. The proposed RVFL-FS method can be fitted into a parallel or distributed computing environment. The RVFL-FS method is employed for the indoor positioning system (IPS) as a regression problem to illustrate the proposed idea. RVFL-FS model works on the idea that instead of using the all hidden nodes, one should select the hidden nodes to generate the robust features. Therefore, RVFL-FS model is computationally efficient and robust as compared to the standard RVFL model. Samal et al. \cite{samal2021modified} proposed a robust non-iterative RVFL, i.e., Added activation function based exponentially enhanced robust RVFL (AAERVFL), wherein trigonometric function based exponentially expanded input vector is connected by a weighted direct link from the input layer to the output layer and a new activation function using a weighted linear combination of two activation functions, i.e., the local sigmoid and global Morlet wavelet function, is introduced. The AAERVFL model is examined with five different non-linear systems and three real-world datasets like electricity price prediction, currency exchange rate prediction, and stock price prediction. The experiments demonstrate the superiority of the AAERVFL model over the standard RVFL.
  
 In industrial processes like the mineral grinding process, noisy data with outliers is acquired due to unavoidable circumstances. Dai et al. \cite{dai2017robust} proposed robust regularised RVFL (RR-RVFL) and its online version as well. Weight calculated from a non-parametric kernel density estimation method is assigned to the empirical error corresponding to each sample. Therefore, the weighting mechanism reduces the negative impact of the outliers over the RR-RVFL model. The experiments over the mineral grinding process demonstrate that the RR-RVFL model is more robust and has better generalization performance than standard RVFL. In RR-RVFL model, weights are assigned using Gaussian kernel function which can be replaced via other efficient functions such as piece-wise continuous function to develop more robust model.
  Predicting the stock market movement is a significant task for future investment.  Chen et al. \cite{chen2021turning} fused the two different algorithms, i.e., RVFL and group method of data handling (GMDH) \cite{mueller1998gmdh}, and proposed RVFL-GMDH model consist of many nice properties such as resist noise/outlier effectively, avoiding the over-fitting problem and has better generalization performance compared to standard RVFL model. The RVFL-GMDH model effectively predicts the turning point of the stock price. Iron and steel-making-based industries are the famous industries in the modern world, and now data-driven models are being employed for the estimation of molten iron quality (MIQ) in these industries.
  
  Cauchy distribution weighted M-estimation-based robust RVFL \cite{zhou2017data} model was developed to estimate the molten iron quality. The training data having outliers affect the modeling capability of standard RVFL, so extra care is needed to handle such data. Therefore, weights are assigned to outliers in the data using Cauchy distribution so that their (outliers) contribution in the modeling process can be identified and the negative influence of these outliers can be reduced. Several techniques such as weighting method  \cite{dai2017robust}, different loss function \cite{zhou2017data}, have been used with RVFL to improve its robustness and generalization performance. The standard RVFL model with the assumption that all the samples are equally important gives equal weights to each sample  for calculating the final parameters but in real world, noisy datasets with outliers are acquired that might have negative impact on the model performance. Therefore, one need to take care of these datasets. In \cite{ashwa2022j}, intuitionistic fuzzy theory is employed that define membership and non-membership function to address above issues with RVFL model and proposed  intuitionistic fuzzy RVFL (IFRVFL) and to check applicability of IFRVFL, it has been employed for diagnosis of Alzheimer's disease. IFRVFL is a robust and binary classifier. Similar works can be seen in  \cite{malikGEIF,NehalGEIF}. Therefore, it should be extended to multiclass problem.  \citet{hazarika20221} proposed robust 1 norm RVFL (1N RVFL) model wherein the optimization problem is solved via newton technique. The 1N RVFL model produce spare outputs and hence, has less number of hidden neurons as compared to standard RVFL model.
  \begin{table}[htbp]
 \centering
    \caption{The summary of activation functions commonly used in RVFL model}
     \label{Table:activation function}
    \resizebox{\textwidth}{!}{
    \begin{tabular}{llllll}
    \hline
S. No.&Activation function& Mathematical formulation \\
\hline 
1&Sigmoid&$\theta(x)=\frac{1}{1+e^{-x}}$\\
\hline
 2&Sign (Signum)&$\theta (x) = 
        \begin{cases}
            -1, ~\quad &x < 0,\\
           0, ~~~&x=0,\\
           1,~~~&x > 0
        \end{cases}
    $\\
\hline
3& Rectified linear unit (Relu)&$\theta(x)=max (0, x)$\\
\hline
4&Sine&$\theta(x)=sin(x)$\\
\hline
5&Radbas&$\theta(x)=e^{-x^{2}}$\\
\hline
6&Hard limit (Hardlim)& 
$\theta (x) = 
        \begin{cases}
            1, ~ \quad x\leq 0\\
           0, ~~\text{otherwise},
        \end{cases}
    $
\\
\hline
8&Tribas&$\theta(x)=max(1-|x|,~ 0)$\\
\hline
9&Hyperbolic tangent (Tanh)&$\theta(x)=\frac{1-e^{-x}}{1+e^{-x}}$\\
\hline
10&Radial basis function (RBF)& $\theta(\mu, \sigma, x)=e^{-\sigma \left \| x-\mu \right \|^{2}}$\\
\hline
11&Multiquadratic&$\theta(\mu,\sigma, x)=\sqrt{\left \| x-\mu \right \|^{2}+\sigma^{2}}$\\
\hline
12&Scaled exponential linear units (Selu)&$\theta(x)=\gamma(max(0,x)+min(0,\alpha(e^{x}-1)))$\\
\hline
\end{tabular}}
\end{table} 
\subsection{Kernelized RVFL models}
\label{Sec:Kernelized RVFL models}
 For training the standard RVFL, one needs to determine the number of enhancement nodes and activation function in advance. Manually determining the optimal range of hidden nodes and the optimal activation function is challenging task. Kernel methods can be used to address aforementioned issues. Chakraavorti et al. \cite{chakravorti2020non} proposed kernel exponentially extended random vector functional link network (KERVFLN) for non-linear system identification.  Expanding the dimension of input vector via trigonometric function and utilizing the expended vector into learning process increase the generalization performance of the RVFL model. Here, the kernel function is used to increase the stability of standard RVFL, and the inputs are extended using a trigonometric function that improves the generalization performance of the KERVFLN model. Based on the analogy of teacher-student interaction mechanism, Zhang and Yang \cite{zhang2020new} proposed RVFL+ and kernel RVFL+ (KRVFL+) model that utilizes the learning using privileged information (LUPI) paradigm into the training process of standard RVFL, and the experiments demonstrate that RVFL+ has better generalization performance compared to standard RVFL model. Moreover, a tight generalization error bound based on the Rademacher complexity is derived for the RVFL+ model and proved the efficiency and effectiveness of the RVFL+ and KRVFL+ models.
 Several machine learning approaches have been developed to classify brain images according to brain abnormalities in the medical domain. Machine learning tools help physicians to make decisions. Nayak et al. \cite{nayak2019application} proposed kernel RVFL (KRVFL) model with a new feature descriptor based on Tsallis entropy and fast curvelet transform to classify brain abnormalities such as brain stroke, degenerative disease, infectious, brain tumor, and normal brain. An efficient hybrid model  \cite{parija2020multi} consists of weighted multi-kernel RVFL network (WMKRVFLN), empirical mode decomposition (EMD) based features, and water cycle algorithm (WCA) was proposed to diagnose and classify the epileptic electroencephalogram (EEG) signals. When number of samples are large enough then Kernel based RVFL models are not applicable over large scale datasets.

\subsection{RVFL with bayesian inference (BI) and other techniques} 
\label{Sec:Improvements in RVFL model based on bayesian inference (BI) and other techniques}
  Scardapane and Wang \cite{scardapane2017bayesian} proposed several alternatives to train standard RVFL by exploiting the Bayesian Inference (BI) framework. In the standard RVFL model, the optimal output weights are generally calculated via (regularized) least square method but here (in \cite{scardapane2017bayesian} ),  probability distribution of the output weights is derived. The Bayesian approach has several advantages, i.e., additional prior information can be introduced in the training process of standard RVFL and the capability of automatically inferring hyper-parameters from given data, etc. Experimental results demonstrate that the Bayesian Inference (BI) based approaches are better than least square approaches (as in standard RVFL). The BI approach to train RVFL model gives a new  area of research to develop robust RVFL model.
 Introducing hybrid regularization term with $L_2$ and $L_1$ norm into standard RVFL, Ye et al. \cite{ye2020hybrid} proposed $L_2-L_1$ RVFL  model that overcomes the stability and sparsity issue of the standard RVFL and gives an iterative algorithm to train $L_2-L_1$ RVFL  model.
  Alalimi et al. \cite{alalimi2021optimized} employed the Spherical Search Optimizer (SSO) algorithm to optimize the RVFL model and named it the SSO-RVFL model. The SSO algorithm improves the parameters of the standard RVFL model. Dai et al. \cite{dai2022incremental} incorporated LUPI paradigm into incremental RVFL (IRVFL) model and proposed IRVFL+ that has strong theoretical foundation. IRVFL+ model has been trained via two approaches. The first one is named IRVFL-I+ that focused on speed of the model and another one is IRVFL-II+ that focused on accuracy.
  Incremental learning based RVFL model solve the problem of constructing an appropriate RVFL model. Here, IRVFL-II+ is computationally expensive as compared to standard RVFL and IRVFL-I+ model. In \cite{jiao2022artificially}, a model for artificially intelligent diagnosis that uses privileged information to learn was proposed to help with ELN differential diagnosis when dealing with single- or dual-modal picture data. In order to create a more effective unmanned aerial vehicle automatic target recognition system, \cite{ribeiro2023random} research suggests two unique machine learning methods, namely Random Vector Functional Link Forests and Extreme Learning Forests. For EEG-based driving fatigue detection, in \cite{zhang2022auto}, an auto-weighting incremental random vector functional link (AWIRVFL) network model that combines incremental learning and online regression prediction. Although AWIRVFL outperformed several deep learning models, its network topology is still shallow, which restricts its feature learning capacity in describing the underlying characteristics of EEG data.
  
In another approach, the RVFL-MO method \cite{abd2021new} optimize the RVFL model via the mayfly optimization (MO) algorithm to predict the performance of solar photovoltaic thermal collector combined with the electrolytic hydrogen production system. Experiments demonstrate that the RVFL-MO model performs better than standard RVFL.
 Elsheikh et al. \cite{elsheikh2021prediction} proposed an enhanced 
 RVFL model with equilibrium optimizer (EO), i.e., RVFL-EO, to predict kerf quality indices during $CO_2$ laser cutting of polymethylmethacrylate (PMMA) sheets. The equilibrium optimizer (EO) algorithm is employed to obtain the parameters of RVFL model that enhance its generalization performance. Several statistical tests were used to compare RVFL-EO with the standard RVFL model, and results indicate the superior performance of  RVFL-EO model.
   A novel classification method \cite{lu2020classification} based on MobileNet and three feed-forward neural networks with random weights, i.e., extreme learning machine (ELM), Schmidt neural network (SNN), and RVFL network, to classify brain magnetic resonance image (MRI) image was proposed. Here, MobileNetv2 is employed to extract features from input brain image, and then the classification task is executed via ELM, SNN, and RVFL models, respectively. The experimental results reveal that the MobileNet-RVFL-CBA method performs better than other proposed MobileNet-ELM-CBA, MobileNet-SNN-CBA methods and compared state-of-the-art methods.
   
 Naive bayes classifier has the capability to handle the mixed data containing categorical and numerical attributes. On the other hand, the one-hot encoding technique is used to deal with categorical features in neural networks. To avoid the use of one-hot encoding with neural network, Ruz and Henriquez  \cite{ruz2019random} proposed a two-stage learning approach based on the RVFL model and Naive Bayes classifier, i.e., RVFL-NB, to handle the mixed data. In the first stage, the Naive Bayes classifier is employed to compute the posterior probabilities for each class, and in the second stage, the RVFL model is trained using as inputs the continuous features and including as additional hidden units the posterior probabilities obtained in the previous step.
 Single hidden layer feed-forward neural networks face a challenge, i.e., how to choose a number of hidden neurons in the hidden layer because this choice leads to underfitting and overfitting phenomena. To address this issue, a non-iterative method  \cite{henriquez2018non} for pruning hidden neurons with random weights was proposed. The pruning method is based on Garson's algorithm and was employed on three neural networks, i.e., single hidden layer neural network with random weights (RWSLFN), RVFL, and ELM, to increase their generalization performance.
 
  Parsimonious RVFL (pRVFL)  \cite{pratama2018parsimonious} model was proposed for the data stream and hence, perform better than the standard RVFL model. pRVFL model has flexible and adaptive working principle wherein its structure is automatically generated and pruned.
In the optical fiber pre-warning system (OFPS), most of the feature extraction methods are quested from the view of the time domain. To address this issue, using multi-level wavelet decomposition,  Wang et al. \cite{wang2018rvfl} extract intrusion signal features of the running, digging, and pick mattock in the frequency domain and then for considering the feature of each intrusion type, the average energy ratio of different frequency bands is obtained. Finally, the RVFL model is trained for the classification and identification of the signal. The results demonstrate that RVFL correctly classifies the different intrusion signals.
El-Said et al. \cite{el2021machine} conducted experiments with four machine learning algorithms, i.e., support vector machine (SVM) \cite{suykens1999least}, K-nearest neighbor (K-NN) \cite{zhang2005k}, sequential minimal optimization regression \cite{candel2010sequential,smola2004tutorial} and RVFL model, to predict the air injection effect on the thermohydraulic performance of shell and tube heat exchanger. The experimental analysis reveals that the RVFL model outperforms compared models with excellent accuracy and better generalization performance. 

\citet{borah2019unconstrained} proposed unconstrained convex minimization based implicit Lagrangian twin RVFL for binary classification (ULTRVFLC) for addressing the overfitting issue in the standard RVFL and hence, has better generalization capability as compared to standard RVFL model. Unlike TWSVM and twin ELM (TELM), in ULTRVFLC, three iterative convergent schemes are employed to make the model computationally efficient. The least-square twin SVM (LSTSVM) \cite{kumar2009least} has been a successful classifier, and it works on original feature space. On the other hand, the RVFL model works on both original and randomized features. Ganaie and Tanveer \cite{ganaie2020lstsvm} proposed a novel LSTSVM model with enhanced features obtained from the pre-trained RVFL model and hence, improved the generalization performance of the base line model.
Prediction of international oil prices has become a hot topic in the field of energy system modeling and analysis. Tang et al. \cite{tang2020multi} proposed a new technique introducing a  multi-scale forecasting methodology with multi-factor search engine data (SED). Incorporating the informative SED, the multi-scale relationship with oil price is explored, and four machine learning models, i.e., ELM, RVFL, linear regression (LR), and backpropagation neural network (BPNN), are employed in this task.
Mary et al. \cite{mary2019random} employed standard RVFL in the image retrival (IR) framework for better performance.
To address the instability issue in the sliding mode control system,
 Zhou and Wu \cite{zhou2021adaptive} proposed an adaptive fuzzy RVFL (FRVFL), wherein self-mapping between fuzzy rules and hidden layers is employed and adaptive rules are also employed to achieve self-adjustment for the output weights. FRVFL model combine RVFL with dynamic fuzzy system to improve its generalization performance. To address the threats issues in android malware detection tools, in \cite{elkabbash2021android}, a novel technique using RVFL model with artificial jellyfish search (JS) optimizer algorithm for selecting the optimal features of android malware datasets, i.e., RVFL+JS, has been proposed. The JS algorithm reduces the redundant and irrelevant features from the data that handle the storage and time complexity issue and hence, improves the generalization performance of the RVFL+JS model. In \cite{zhou2023fabric}, ResNet18 was used to extract wrinkle image features, and an RVFL algorithm optimised by the TSA algorithm based on logistic maps and opposition based learning was proposed for evaluating fabric wrinkle level. This was done to address the issues of low accuracy and low efficiency in evaluating the wrinkle degree by visual perception and the shortcomings of the current artificial neural networks in evaluating the wrinkle level. However, the amount of fabric samples at some levels was insufficient, and just a few different fabric types were included in the study's fabric samples. Label distribution learning (LDL), as opposed to multi-label learning (MLL), can reflect the importance of pertinent labels in samples, which is why many LDL studies have lately been appearing. In \cite{huang2022online}, an unique LDL framework based on RVFL is proposed, which can efficiently and precisely handle the live data stream.
 
\subsection{Imbalance learning based on RVFL model}
\label{Sec:Imbalance learning based on RVFL model}
The class imbalance problem occurs when one class has small samples compared to other classes. Standard RVFL is not capable of handling imbalance data. Cao et al. \cite{cao2020improved} proposed improved fuzziness based RVFL (IF-RVFL) where synthetic minority over-sampling technique (SMOTE) \cite{chawla2002smote} is combined with fuzziness based RVFL model \cite{cao2017fuzziness}. The experiments on real-life liver disease dataset demonstrate that the IF-RVFL model performs better than standard RVFL and F-RVFL model.
 To diagnose the power quality disturbance (PQD), Sahani and Dash \cite{sahani2019fpga} proposed class-specific weighted RVFL (CSWRVFL). Here, a novel signal decomposition technique- reduced sample empirical mode decomposition- is proposed to extract the highly correlated monocomponent mode of oscillations. Hilbert transforms (HT) extracted the two effective power quality indices are extracted from Hilbert transforms (HT). Finally, the combined framework RSHHT-CSWRVFL is applied for online monitoring the power quality disturbances (PQDs) with better classification accuracy.\\
  
\subsection{Multi-label classification based on RVFL model}
\label{Sec:Multi-label classification based on RVFL model}
 In a traditional classification problem, each sample is associated with only one target label from a set of labels. However, each sample can be related to more than one label in multi-label classification problems. There are several fields, i.e., medical diagnosis, music categorization, etc., wherein multi-label data is produced. Chauhan and Tiwari \cite{chauhan2021randomized} extended the standard RVFL for the multi-label task. In \cite{chauhan2021randomized}, randomization based neural networks, i.e., multi-label RVFL (ML-RVFL), multi-label kernelized RVFL (ML-KRVFL), multi-label broad learning system (ML-BLS), and multi-label fuzzy BLS (ML-FBLS) were proposed to handle the multi-label classification problems. Here, the ML-KRVFL model performs better than other compared models.

Table \ref{tab:Summary of shallow RVFL} shows the summary of the shallow RVFL model and its variants. The table gives a highlight of the journey of shallow RVFL in tabular form with variants name, activation functions, hyper-parameter optimization method/solution and finally their applications. In summary, researchers have developed several variants of RVFL model using various methods such as different kind of initialization techniques \cite{zhang2016comprehensive,zhang2019unsupervised}, kernel methods \cite{zhang2020new,chakravorti2020non}, manifold learning \cite{guo2021assessing,ganaie2020minimum}, fuzzy theory \cite{ashwa2022j} and so on. Kernel based RVFL models \cite{parija2020multi,nayak2019application} are robust, stable and have  better generalization performance than standard RVFL model. There is no need to choose activation functions and hidden nodes in kernel based RVFL models, however, these models are not suitable for large scale datasets (when $N$ is large enough). Standard RVFL doesn't consider geometric informations of the data while calculating the output weights whereas, using these kind of informations, RVFL variants with manifold learning theory have better generalization than standard RVFL.


\tiny
\begin{landscape}
\begin{longtable}[t]{p{0.02\paperheight}p{0.08\paperheight}p{0.2\paperheight}p{0.1\paperheight}p{0.17\paperheight}p{0.15\paperheight}}
    \caption{The summary of shallow RVFL models}
     \label{tab:Summary of shallow RVFL}
\\
\hline
Year&Literature&Model description&Activation function&Hyper-parameter optimization or Solution&Application  \\
\hline
2022&\citet{ganaieminimum}&MVRVFL+&Relu&Closed form&Classification\\
\hline
2022&\citet{ashwa2022j}& Intuitionistic fuzzy RVFL (IFRVFL)&Selu, relu, sigmoid, sin, hardlim, tribas, radbas, sign function&Closed form&Alzheimer's disease diagnosis\\
\hline
2022&\citet{hazarika20221}&1-norm RVFL (1N RVFL)&Sine, relu&Newton technique&Classification problem\\
\hline
2021& \citet{samal2021modified}& A non-iterative robust AAERVFL&Sigmoid, global
morlet wavelet functions &Closed form &Nonlinear system identification\\
 \hline
2021& \citet{parija2021deep}& Minimum variance based kernel RVFL (MVKRVFL) &-&  Closed form &Epileptic EEG signal classification\\
\hline 
 2021&\citet{chen2021turning}&  Group method of data handling (GMDH) based RVFL (RVFL-GMDH)&Sigmoid& Iterative method  &Stock price prediction\\
\hline
2021&\citet{zhou2021adaptive}& Adaptive fuzzy RVFL (FRVFL)  &-&Iterative method&Slide mode control for manipulators\\ 
\hline
2021&\citet{zhang2021reinforced}& Reinforced fuzzy clustering-based rule model (RFCRM) &-&Iterative method&Regression problem\\
\hline
2021&\citet{elkabbash2021android}&RVFL+JS&-&Jellyfish search optimizer&Android malware classification\\

\hline
2021&\citet{guo2021assessing}&Sparse Laplacian regularized RVFL (SLapRVFL)  &- &Iterative method &Assessing dry weight of hemodialysis patients\\
 \hline
2021& \citet{alalimi2021optimized}& Spherical search optimizer based RVFL (SSO-RVFL)&Tribas, sign, hardlim, radbas, sin, sig &Spherical search optimization algorithm &Prediction of oil production in China\\
\hline
2021& \citet{gao2021walk}&Walk-forward EWT based RVFL (WFEWT-RVFL)&-&Closed form &Time series forecasting\\
\hline
2021& \citet{tanveer2021ensemble}&TBSVM-FL, TWKSVC-FL, LSTWKSVC-FL, RELSTSVM-FL, LSTSVM-FL&Radbas&Closed form (LS)& Classification problem\\
\hline
2021&  \citet{abd2021new}&Mayfly optimization (MO) algorithm based RVFL (RVFL-MO) &Sign, hardlim, sig, tribas, radbas&Mayfly optimization algorithm& solar photovoltaic thermal collector combined with electrolytic hydrogen production system\\ 
\hline
2021& \citet{elsheikh2021prediction}&Equilibrium optimizer based RVFL (RVFL-EO)&Sign, hardlim, sig, tribas, radbas&Equilibrium optimization algorithm &Laser cutting parameters for polymethylmethacrylate sheets\\  
\hline
 2021& \citet{zayed2021predicting}&RVFL-CHOA&Sign, tribas, sigmoid, hardlim, radbas& Chimp optimization algorithm (CHOA) &Solar power forecasting\\
 \hline
2021&\citet{ganaie2021co}&Co-trained RVFL (coRVFL)&-&Closed form&Classification problem\\
 \hline
2021& \citet{dash2021short}&Empirical wavelet transform based robust minimum variance RVFL (EWT-RRVFLN)&Local sigmoid function, global morlet wavelet&Closed form (LS)&Short term solar power forecasting\\
\hline
2021& \citet{gao2021walk}&Walk forward empirical wavelet transformation based RVFL (WFEWT-RVFL)&-&-& Time series forecasting \\
\hline 
 2021&\citet{dai2022incremental}&Incremental learning paradigm with privileged information for random vector functional-link networks: IRVFL+&Sigmoid, sine, triangular, radial& Incremental learning&Classification and regression problems\\
\hline
2020& \citet{ganaie2020lstsvm}& LSTSVM classifier with enhanced features from pre-trained RVFL&Radbas&Closed form& Classification problem\\
\hline
2020&\citet{ganaie2020minimum}&Total-Var-RVFL, Class-Var-RVFL&Relu&Closed form&Classification problem\\
\hline
2020& \citet{hazarika2020modelling}&Wavelet-coupled RVFL (WCRVFL) network&Relu, sigmoid&Closed form (LS) &COVID-19 cases forecasting\\
\hline
 2020& \citet{dudek2020direct}& Standard RVFL&Sigmoid&Closed form (LS) &Regression problem \\
\hline
 2020&\citet{chakravorti2020non}& Exponentially extended RVFL network (ERVFLN), Kernel ERVFLN (KERVFLN)&Tanh(.)& Closed form (LS)      &Nonlinear system identification\\
\hline

 2020&\citet{cao2020study}&- &Sigmoid, radial basis function (RBF), sine&    &Classification problem\\
\hline
2020& \citet{lu2020classification}&  MobileNet-RVFL-CBA, MobileNet-ELM-CBA and MobileNet-SNN-CBA&-& Closed form (LS)    & Brain MRI image classification problem\\
 \hline
2020&\citet{rasheed2020respiratory}& Standard RVFL&Sine, hardlim, sigmoid, tribas, radbas, sign&Closed form (LS) &Respiratory motion prediction\\
\hline
2020& \citet{zhang2020new} & RVFL+, kernel RVFL+ & Sigmoid, sine, hardlim, triangular basis function (TBF), radial basis function (RBF)&Closed form (LS) &Classification and regression problems\\
\hline

2020&\citet{ye2020hybrid} & $L_{2}-L_{1}$-RVFL &Sigmoid&Iterative method & Classification Problem\\
\hline
2020& \citet{abd2020utilization}&MPA-RVFL& Sigmoid, sine, hardlim, tribas, radbas& Marine predators algorithm (MPA) & Tensile behavior prediction\\
\hline
2020& \citet{ESSA2020322}&RVFL-AEO&Sign, tribas, sigmoid, hardlim, radbas& Artificial ecosystem-based optimization (AEO) algorithm &Forecasting power consumption and water productivity of seawater\\
\hline
 2020&\citet{sharshir2020enhancing}&Firefly algorithm based RVFL (FA-RVFL)&Sigmoid, radbas, hardlim, sine, sign, tribas&Firefly algorithm&Thermal performance and modeling prediction of developed pyramid solar\\
 \hline
2020&\citet{pan2020random}&  Jaya-RVFL &-& Jaya algorithm&Transient stability assessment of power systems\\
\hline
2020& \citet{parija2020multi}& Empirical mode decomposition (EMD) and water cycle algorithm (WCA) based weighted multi-kernel RVFL network (WMKRVFLN) (WCA-EMD-WMKRVFLN) &-& Closed form (LS)& Epileptic EEG signal classification\\
\hline
2019& \citet{hussein2019new}&Moth search algorithm based RVFL (MSA-RVFL)&-&Closed form (LS)& Water quality analysis \\
\hline
 2019&\citet{zhang2019unsupervised}& Sparse auto encoder with $l_1$ norm based RVFL (SP-RVFL) &Sigmoid&Closed form (LS)&Classification problem\\

\hline
2019& \citet{nayak2019application}&kernel RVFL (KRVFL) &-&Closed form (LS) & Brain abnormalities detection\\
\hline
2019& \citet{zhou2019data}&Improved orthogonal incremental RVFL (I-OI-RVFL)&Sigmoid&Iterative method&Data modeling \\
\hline
 2019& \citet{ruz2019random}& Naive Bayes classifier based RVFL (RVFL-NB) &Sigmoid&closed form&Classification problem of mixed data\\
 \hline
2019&\citet{cao2019initial}&Standard RVFL & Sigmoid &Closed form (LS)   &Classification problem\\
\hline
2019&\citet{bisoi2019modes}&Variational mode decomposition based RVFL (VMD-RVFL)&Tanh&Closed form (LS)& Crude oil price forecasting\\
\hline
2019& \citet{borah2019unconstrained}&Unconstrained convex minimization based implicit Lagrangian twin RVFL for binary classification (ULTRVFLC)&Multiquadratic& Newton-Armijo stepsize method&  Classification problem\\
\hline
 2019& \citet{sahani2019fpga}& Class-specific weighted RVFL (CSWRVFL) &Tanh & Closed form (LS)& Power quality disturbances\\
  \hline
 2019& \citet{fan2019image}& Random subspace fisher linear discriminant (FLD) based RVFL&-& Closed form (LS)   &Image steganalysis\\
 \hline
2018& \citet{wang2018rvfl}&Standard RVFL  &-& Closed form (LS)      &Optical fiber pre-warning system\\
\hline
 2018&\citet{pratama2018parsimonious}&Parsimonious RVFL (pRVFL)&- &FWGRLS method &Data stream\\
\hline
2018& \citet{henriquez2018non}&Neural networks with random weights (NNRW) &Sigmoid&Closed form (LS)&Regression and classification problem \\
\hline
2018& \citet{vukovic2018comprehensive}& Orthogonal polynomial expanded RVFL (OPE-RVFL) &Tansig, logsig, tribas & Closed form (ridge regression/ Moore-Penrose pseudoinverse) & Regression problem\\
\hline
2018& \citet{dash2018indian}&Standard RVFL, regularized online sequential network (ROS-RVFL)&Radbas&Closed form (LS)&Indian summer monsoon rainfall prediction\\
\hline
 2018&\citet{nhabangue2018wind}&Empirical mode decomposition based improved RVFL (EMD-IRVFL) &-&Closed form (LS)&Wind speed forecasting\\
\hline
2017& \citet{cui2017received}&   Feature selection based RVFL (RVFL-FS)&-&Alternative direction method of multiplier (ADMM) algorithm &Fingerprinting based indoor positioning system \\ 
\hline
2017& \citet{scardapane2017bayesian}&  Bayesian inference RVFL (B-RVFL)&-& Bayesian inference algorithm     &Classification problem\\
\hline
2017& \citet{cao2017impact}&Standard RVFL &Sigmoid& Closed form (LS)      & Regression problem\\
\hline
2017&\citet{dai2017robust}& Robust regularized RVFLN (RR-RVFLN), online robust regularized RVFL (ORR-RVFLN) &-&Closed form (LS)/ Iterative method&Industrial application\\
\hline
 2017&\citet{zhou2017data}& Cauchy distribution weighted M-estimation-based robust RVFL (Cauchy-M-RVFLN) &Sigmoid&Iterative method &Blast furnace iron-making process\\
\hline
2017& \citet{xu2017kernel}&Kernel RVFL (K-RVFL)& -& Closed form (LS)&Thermal process\\
\hline
 2017&\citet{dai2017healthy}& Manifold learning based RVFL        &-&Closed form (LS) & Analysis of alzheimer’s disease\\
\hline
2016& \citet{zhang2016multivariable}&Multivariable incremental RVFL (M-I-RVFLN)&-&Iterative method&Molten iron quality prediction\\
\hline
2016& \citet{zhang2016comprehensive}&Standard RVFL&Sigmoid, sine, hardlim, tribas, radbas, sign&Closed form (ridge regression/ Moore-Penrose pseudoinverse)&Classification problem\\
 \hline
2015&\citet{zhou2015multivariable}&Online sequential RVFL (OS-RVFL)&-&Iterative method&Molten iron quality prediction\\
\hline
2015& \citet{ren2015detecting}& Standard RVFL & Logistic sigmoid&Closed form (LS)&Wind power ramp detection\\


 \hline

  \end{longtable}
  
\end{landscape}
\normalsize


\section{Semi-supervised methods based on RVFL model}
\label{Sec:Semi-supervised methods based on RVFL model}
There are datasets in which small number of samples are labeled in many applications. RVFL variants have been successfully employed in diverse domains, i.e., classification and regression, etc. But there is very little research for solving semi-supervised learning problems with the RVFL model. Table \ref{tab:The summary of semi-supervised RVFL models} summarizes the RVFL models developed for the semi-supervised learning. Peng et al. \cite{peng2020joint} proposed a joint optimized semi-supervised RVFL model, i.e., JOSRVFL, in which a novel approach is used to optimize the objective function of the JOSRVFL model. There are many techniques in machine learning to improve the generalization performance of a model, and fuzzy theory is one of them. In \cite{cao2017fuzziness}, a novel fuzziness-based RVFL model has been proposed for a semi-supervised learning problem. Inspired by transductive SVM \cite{wang2007transductive}, Scardapane et al. \cite{scardapane2016semi} proposed a transductive RVFL (TR-RVFL) model that defines box-constrained quadratic (BCQ) problem solvable in polynomial time. The TR-RVFL model performs better than many state-of-art algorithms based on the manifold regularization (MR) theory. In \cite{xie2020distributed}, two algorithms, i.e., horizontally distributed semi-supervised learning (HDSSL) and vertically DSSL, were proposed. Both algorithms are based on the RVFL model and alternating direction method of multipliers (ADMM) strategy. The HDSSL and VDSSL algorithms solve DSSL problems with horizontally and vertically partitioned data, respectively. Therefore, the RVFL model performs well in semi-supervised problems.
\begin{table}[htbp]
    \centering
    \caption{The summary of semi-supervised RVFL models}
    \label{tab:The summary of semi-supervised RVFL models}
    \resizebox{\textwidth}{!}{
    \begin{tabular}{llllll}
    \hline
Year&Literature&Model description&Activation function&Hyper-parameter optimization or Solution&Application \\
\hline 
2020&\citet{peng2020joint}& JOSRVFL and JOSELM&-&Iterative method&Classification problem\\
\hline
 2020&\citet{xie2020distributed}&Horizontally distributed  &-&Iterative method& Classification problem\\
 &&semi-supervised learning (HDSSL), &&&\\
 &&vertically DSSL (VDSSL) &&&\\
\hline
2017& \citet{cao2017fuzziness}&F-RVFL&Sigmoid&Closed form&Classification problem\\
\hline
2016& \citet{scardapane2016semi}&TR-RVFL&Sigmoid&Closed form&Classification problem\\
\hline
\end{tabular}}
\end{table}

\section{Clustering methods based on RVFL model}
\label{Sec:Clustering methods based on RVFL model}
Clustering is an unsupervised learning problem where samples are categorized into clusters (groups) based on their similarities. In the literature, there are various clustering techniques, i.e., point-based clustering methods \cite{jain1988algorithms} and plane based clustering \cite{tanveer2021pinball,richhariya2020least}, etc. 
\citet{zhang2021unsupervised} proposed an unsupervised discriminative RVFL (UDRVFL) model for the clustering problem. To capture the local information within data, the local manifold learning concept has been used while global biased knowledge of the data has also been considered so that data can be clustered in an optimal manner.

\section{Ensemble frameworks based on RVFL model}
\label{Sec:Ensemble frameworks based on RVFL model}
Ensemble learning utilizes multiple learning algorithms, which are named base learners. The performance of a single RVFL model is often unstable because of the random nature of its hidden features. To improve the model's stability and performance, it can be beneficial to average the outputs from multiple RVFL models that each have different hidden features. This approach is commonly known as ensemble learning. Ensembles of randomized models, such as random forests, are effective in reducing the variance resulting from the random feature space. Thus, researchers working with RVFL models have explored the development of ensemble RVFLs to improve the model's stability and performance. In general, there are two steps in ensemble learning. First, a pool of base learners is constructed parallel or sequential. Second, the base learners are combined for decisions according to some rules. Therefore, the ensemble RVFL method either trains multiple RVFLs or utilizes a meta-RVFL to connect the outputs of the base learners. Table \ref{tab:Summary of ensemble RVFL} summarizes the representative literature about the ensemble RVFL models.

\subsection{Ensemble RVFL-based on bagging}
Bagging is short for bootstrap aggregating, which trains the base leaner using a subset from the training data. The subsets are drawn randomly with replacement \cite{zhou2021ensemble}. For instance, the base learners of a rotated forest are replaced by RVFLs for classification problems \cite{maliknovel}. Bagging generates different subsets with different features to train the RVFL with different structures \cite{yu2021investigation}. 
\subsection{Ensemble RVFL-based on boosting}
Boosting constructs the ensemble RVFL incrementally by paying more attention to the samples which are not correctly learned by the base RVFL. In \cite{zhang2019robust}, Maximum Relevance Minimum Redundancy is utilized to select features, and the ensemble RVFL is constructed using the Adaboost scheme.
\subsection{Ensemble RVFL-based on stacking}
Stacking refers to training a meta-learner that works on the outputs from all base learners. In \cite{tahir2020novel}, a meta-RVFL is trained to combine the results from all the base RVFL networks with different activation functions. In \cite{qiu2018ensemblecrude}, an individual RVFL network is trained to forecast each sub-series generated by the decomposition, and an incremental RVFL is introduced to aggregate all forecasts. In \cite{qiu2018ensembleload}, the short-term load is decomposed by a two-stage decomposition, and an incremental RVFL aggregates the forecasts from each sub-series RVFL. 
\subsection{Ensemble RVFL-based on decomposition}
\begin{figure*}[htbp]
	\centering
	\includegraphics[width=\textwidth]{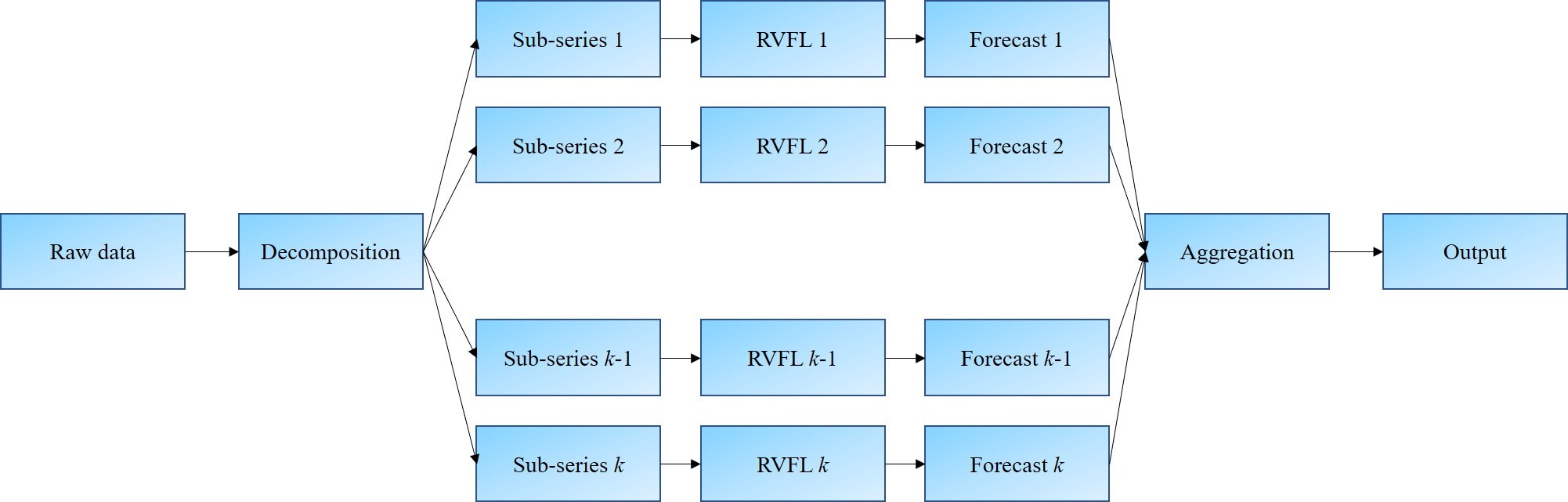}
	\caption{The architecture of decomposition-based ensemble RVFL.}
	\label{fig:dec ensemble RVFL}
\end{figure*}

Another branch of ensemble RVFL is the decomposition-based ensemble framework shown in Figure \ref{fig:dec ensemble RVFL}. The time series is decomposed into sub-series carrying different frequencies and each sub-series is modeled by an individual RVFL network. Finally, the aggregation of all forecasts is the output. There are many mature signal decomposition algorithms, such as the empirical mode decomposition (EMD) \cite{huang1998empirical}, bi-variate empirical mode decomposition (BEMD) \cite{rilling2007bivariate}, ensemble empirical mode decomposition (EEMD) \cite{wu2009ensemble}, complete ensemble empirical mode decomposition (CEEMD) \cite{torres2011complete}, hybrid decomposition \cite{qiu2019fusion,cheng2019classification} and other algorithms \cite{gilles2013empirical,dragomiretskiy2013variational}. For the decomposition block, there are many hybrid ensemble RVFL with signal decomposition algorithms, such as EMD \cite{qiu2016electricity,qiu2018ensemblecrude,zhang2020novel}, EEMD \cite{tang2018randomized,tang2018non,li2018travel,sun2021privileged}, CEEMD \cite{qiu2017short,wu2020hybrid,wu2020daily} and two-stage decomposition \cite{qiu2018ensembleload}.  In \cite{qiu2018ensembleload}, the discrete wavelet transform (DWT) is utilized to decompose the modes generated by EMD into sub-series and an incremental RVFL is trained to aggregate the forecasts from all sub-series. Technical indicators are also utilized to augment the decomposed features for stock forecasting and stock trend classification \cite{qiu2019fusion,cheng2019classification}.

\subsection{Ensemble weights}
Determination of the ensemble weights is crucial for the final performance. If large weights are assigned to a bad base RVFL, it is a disaster for the overall performance. Most of the ensemble RVFL employs the equal-weight scheme \cite{qiu2016electricity,qiu2017short,zhang2020novel,tang2018randomized,tang2018non,li2018travel,sun2021privileged,wu2020hybrid}. Besides the equal-weight scheme, different algorithms are proposed and applied to learn such unequal ensemble weights. For instance, evolutionary algorithms are adopted to learn, and ensemble weights, as well as the base learners' parameters \cite{yu2021investigation,lu2018prediction,lian2018constructing}. A negative correlation learning strategy is utilized to learn the ensemble weights in \cite{lu2018ensemble,miskony2017randomized}.
 
\subsection{Diverse model pool}
Besides the above ensemble RVFL, whose base learners are all RVFL networks, some researchers utilize multiple machine learning models, including RVFL, to construct the ensemble pool, which increases the models' diversity. For instance, ELM, RVFL, and Schmidt neural networks (SNNs) are trained on the same features generated from a DL model in \cite{lu2018ensemble}. Finally, the majority voting mechanism combines the outputs from these three neural networks. In \cite{mesquita2018building}, the successive projections algorithm is utilized to build ensemble ELM, RVFL, and feedforward networks with random weights. In \cite{tang2018randomized}, ELM, BPNN, and RVFL are employed to forecast each sub-series after decomposition. In \cite{ren2019hybrid}, the model pool consists of RVFL and ELM, which are trained in an offline fashion first, and only a subset of them is randomly selected for online updating. In \cite{xia2019data}, the fast Fourier transform and Relief algorithms extract features for ensemble ELM and RVFL.

\subsection{Other diversity strategies}
Besides the diversity strategies of typical ensemble learning, novel diversity strategies specifically designed for RVFL networks are investigated \cite{yu2021investigation,liu2020ensemble}. In \cite{liu2020ensemble}, different RVFLs' enhancement features are initialized according to different distributions. Five novel diversity strategies, such as data quantity diversity, sampling interval diversity, parameter diversity, ensemble number diversity, and ensemble method diversity, are proposed and investigated in \cite{yu2021investigation}. \citet{zhang2017benchmarking} propose a novel and efficient strategy to increase the diversity of ensemble RVFL. A single RVFL is trained, and its hyper-parameters are optimized by cross-validation. Then the other RVFLs' hyper-parameters are generated by adding a noisy deviation to the optimal value.
\subsection{Others}
RVFL also helps split the dataset into subsets, and each subset is learned by the oblique decision tree \cite{katuwal2018enhancing}. In \cite{shi2018cascaded}, a cascaded ensemble RVFL where the shallow layers' RVFL generate predicted values for the successive layers. \citet{9851673} proposed a novel ensemble model (en-efRVFL) which has extended features based RVFL model as base model and the output of each base model is integrated by averaging method. The en-efRVF model has three kinds of features, i.e., original features, supervised randomized features (newly generated) and unsupervised randomized features and therefore generates more accurate and diverse base model in the ensemble. In \cite{sharma2022conv}, conv-eRVFL model combines the CNN model with ensemble RVFL model and implement it for the diagnosis of Alzheimer’s Disease. An ensemble of RVFL models is fed with the features that an eight-layer trained CNN derives from multiple layers. The s-membership fuzzy function is incorporated into the RVFL network as an activation function to help deal with outliers. In order to reach a judgement, the outputs of all the bespoke RVFL classifiers are averaged and supplied to the RVFL classifier. 

\begin{table}[htbp]
    \centering
    \caption{Summary of ensemble RVFL models}
    \label{tab:Summary of ensemble RVFL}
    \resizebox{\textwidth}{!}{
    \begin{tabular}{llllll}
     \hline
         Year&Literature&Model pool&Diversity strategy&Ensemble strategy&Application  \\
         \hline
        2022&\citet{9851673}&Extended feature RVFL (efRVFL) &-&Averaging&Classification Problem\\
         \hline
         2021&\citet{yu2021investigation}&RVFL&\makecell[l]{Data quantity diversity\\Sampling interval diversity\\Parameter diversity\\Ensemble number diversity\\Ensemble method diversity}&\makecell[l]{Averaging\\Adaboost\\RVFL}&Forecasting\\
          \hline
          2021&Lu et al. \cite{lu2021cerebral}&RVFL,ELM and Schmidt NN&-&Majority voting mechanism&Cerebral microbleed diagnosis\\
          \hline
          2021&\citet{hu2021short}&RVFL&Solutions' angle&Evolutionary algorithm&Forecasting\\
          \hline
       
        2021&\citet{maliknovel}&RVFL&Bagging&Rotated forest&Classification  \\
        \hline
       2020&Liu et al. \cite{liu2020ensemble}&RVFL&Different distributions&Majority voting mechanism&Classification  \\
       \hline
       2020&Tahir et al. \cite{tahir2020novel}&RVFL&Stacking&RVFL&Multichannel fall detection  \\
       \hline
       2019&Chen et al. \cite{chen2019selective}&RVFL&Bootstrap&Game theory&Classification  \\
       \hline
      2019&\citet{musikawan2019parallelized}&RVFL&Metaheuristic algorithm&Liear regression&Regression  \\
      \hline
    2019&\citet{xia2019data}&RVFL and ELM&Diversity strategy&Ensemble strategy&IGBT open-circuit fault diagnosis  \\
       \hline
       2018&\citet{shi2018cascaded}&RVFL&Diversity strategy&Ensemble strategy&Parkinson's disease diagnosis  \\
       \hline
       2018&\citet{mesquita2018building}&RVFL, ELM and randomized feedforward NN&Successive Projections Algorithm&Successive Projections Algorithm&Regression  \\
       \hline
      2018&\citet{lu2018prediction}&RVFL&De-correlation&Negative correlation learning&Forecasting  \\
      \hline
      2018&\citet{katuwal2018enhancing}&Oblique decision tree&The RVFL splits the data into subsets.&&Classification  \\
      
       \hline
        2018&\makecell[l]{\citet{li2018travel}\\\citet{qiu2016electricity}\\\citet{sun2021privileged}\\\citet{tang2018non,tang2018randomized}\\\citet{zhang2020novel}}&RVFL&Signal decomposition&Summation&Forecasting  \\
       \hline
        2017&\citet{qiu2017short}&Kernel ridge regression&Signal decomposition&RVFL&Forecasting  \\

      \hline
    2017&\citet{miskony2017randomized}&RVFL&De-correlation&Negative correlation learning&Prediction interval  \\
    
    \hline
    \end{tabular}}
\end{table}
\section{Deep architectures based on RVFL model}
\label{Sec:Deep architectures based on RVFL model}
The success of deep learning (DL) is based on the hierarchical representations of the raw data \cite{alzubaidi2021review}. DL stacks multiple hidden layers and optimizes the weights using any variants of the back-propagation algorithm. With the help of the deep architecture, the DL can extract multi-scale features automatically. Inspired by the idea of DL, deep RVFL with multiple enhancement layers has been proposed \cite{shi2021random}. 

Table \ref{tab:Summary of deep RVFL} summarizes the representative literature about different deep RVFL networks. The main distinction among them is the utilization of direct links. Some literature only utilizes the direct link to connect input layer to output layer to assist in learning the linear patterns \cite{zhang2016visual,henriquez2018twitter}. Some literature utilizes direct links to connect all hidden layers and output layers \citep{gao2021random,gao2022inpatient,shi2022weighting,shi2021random,CHENG2021107826}. Therefore, the raw features are utilized to provide clean information to each level's representation. Furthermore, \citet{katuwal2019stacked} proposed dense connections of all hidden layers. Another characteristic of the deep RVFLs is the number of output layers. Multiple output layers benefit from the ensemble learning and improving the performance \citep{gao2021random,gao2022inpatient,shi2022weighting,shi2021random,CHENG2021107826}. 

The main research problem of deep RVFL is the architecture design. This section reviews the state-of-the-art deep RVFL' architectures in detail. The state-of-the-art deep RVFL architectures are shown in Figure \ref{fig:edRVFL}. These architectures can be classified into three categories, the stacked deep RVFL, the hybrid deep RVFL, and the ensemble deep RVFL (edRVFL).  

\begin{table}[htbp]
    \centering
    \caption{Summary of deep RVFL models}
    \label{tab:Summary of deep RVFL}
    \resizebox{\textwidth}{!}{
    \begin{tabular}{llllll}
    \hline
         Year&Literature&Direct link&Output layer&Activation function&Application  \\
         \hline
         2022&\citet{hu2022edrvflR}
         &Random skip connections&Multiple output layers&Four activation functions &Regression \\
         \hline
         2022&\citet{du2022time}
         &From input to each hidden and output layer&Dynamic ensemble&$sigmoid$ &Forecasting \\
         \hline
         2022&\citet{yu2022selective}
         &From input to each output layer&Multiple output layers&Na &Landslide displacement prediction \\
         \hline
         2022&\citet{hu2022automated}
         &Automatic search&Multiple output layers&Automatic search &Classification \\
         \hline
         2022&\citet{shi2022jointly}
         &From input to each hidden and output layer&\textbf{Multiple output layers}&$Sigmoid$ &Semi-supervised classification \\
         \hline
         2022&\citet{shi2022weighting}&From input to each hidden and output layer&\textbf{Multiple output layers}&$Sigmoid$ &Forecasting \\
         \hline
         2022&\citet{gao2022inpatient}&From input to each hidden and output layer&\textbf{Multiple output layers}&$Sigmoid$ &Forecasting \\
         \hline
         2022&\citet{ganaie2022ensemble}&From input to each hidden and output layer&Multiple output layers&Different activations &Diagnosis of Alzheimer disease\\
         \hline
         2022&\citet{malik2022GE}&From input to each hidden and output layer&Multiple output layers&Different activations&Diagnosis of Alzheimer disease\\
         \hline
         2021&\citet{sharma2021faf}&-&-&s-fuzzy activation function&Diagnosis of Alzheimer disease\\
            \hline
         2021&\citet{shi2021random}&From input to each hidden and output layer&Multiple output layers&$Sigmoid$ &Classification  \\
            \hline
            2021&\citet{CHENG2021107826}&From input to each hidden and output layer&Multiple output layers&Five activation functions &Time series classification  \\
            \hline
            2021&\citet{dai2021sar}&Without direct link&Last hidden layer's features &Sign&SAR target recognition\\
    \hline
            2019&\citet{katuwal2019stacked}&Densely connected to hidden layers &Last hidden layer's features&$Sigmoid$ &Classification  \\
            \hline
            2019&\citet{katuwal2019stacked}&From input to output layer&Last hidden layer's features&$Sigmoid$ &Classification  \\
            \hline
            2018&\citet{henriquez2018twitter}&From input to output layer&Last hidden layer's features&$Sigmoid$ &Sentiment classification  \\
            \hline
          2017&\citet{zhang2016visual}&From input to output layer&Last hidden layer's features&$Relu$ &Visual tracking  \\
          \hline

    \hline
    \end{tabular}}
\end{table}

\begin{figure*}
\subcaptionbox{sdRVFL \cite{katuwal2019stacked}}{
	\includegraphics[width=.85\textwidth]{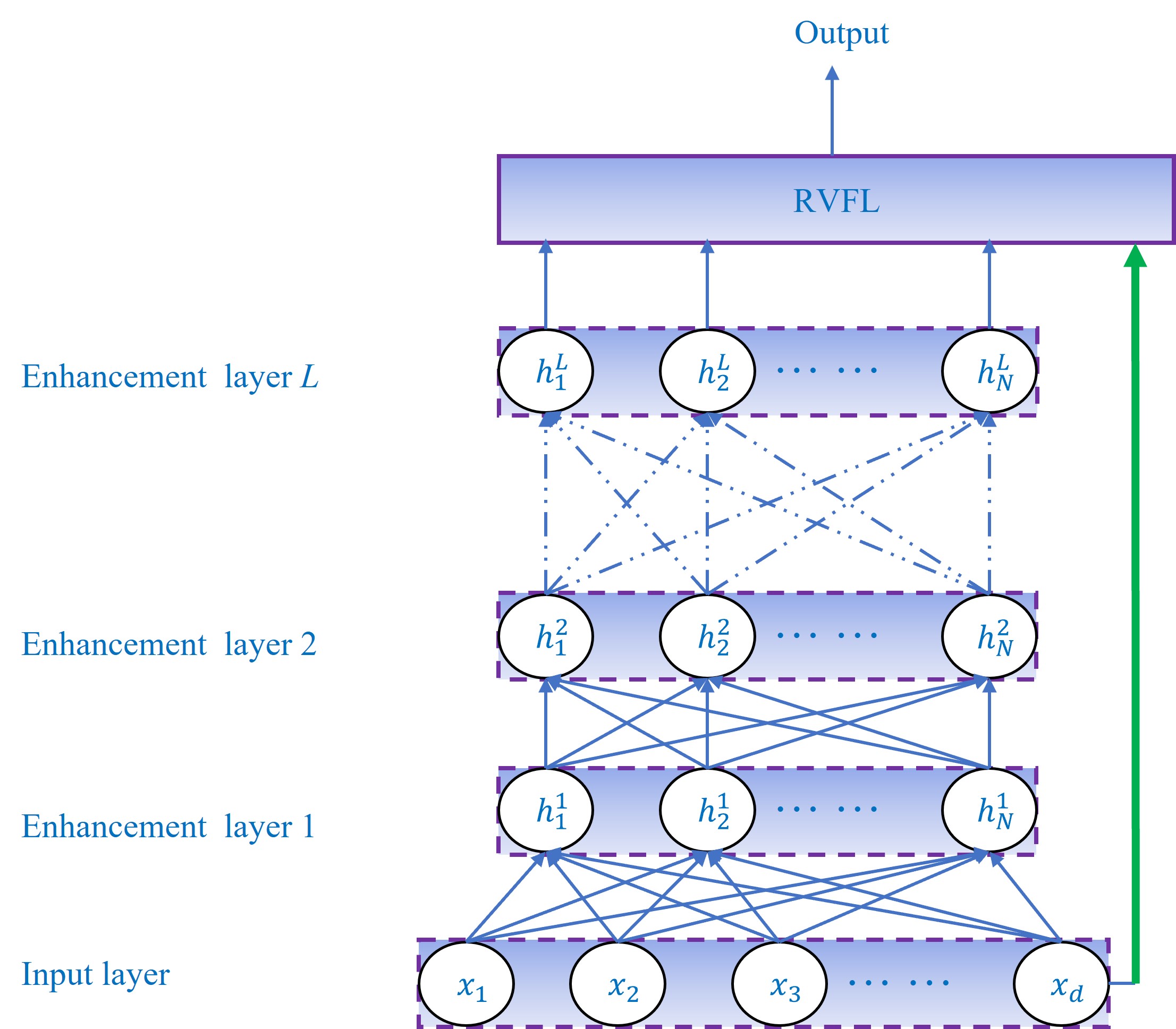}}
	\label{fig:sdRVFL}

\subcaptionbox{sdRVFL(dense) \cite{katuwal2019stacked}}{
	\includegraphics[width=.85\textwidth]{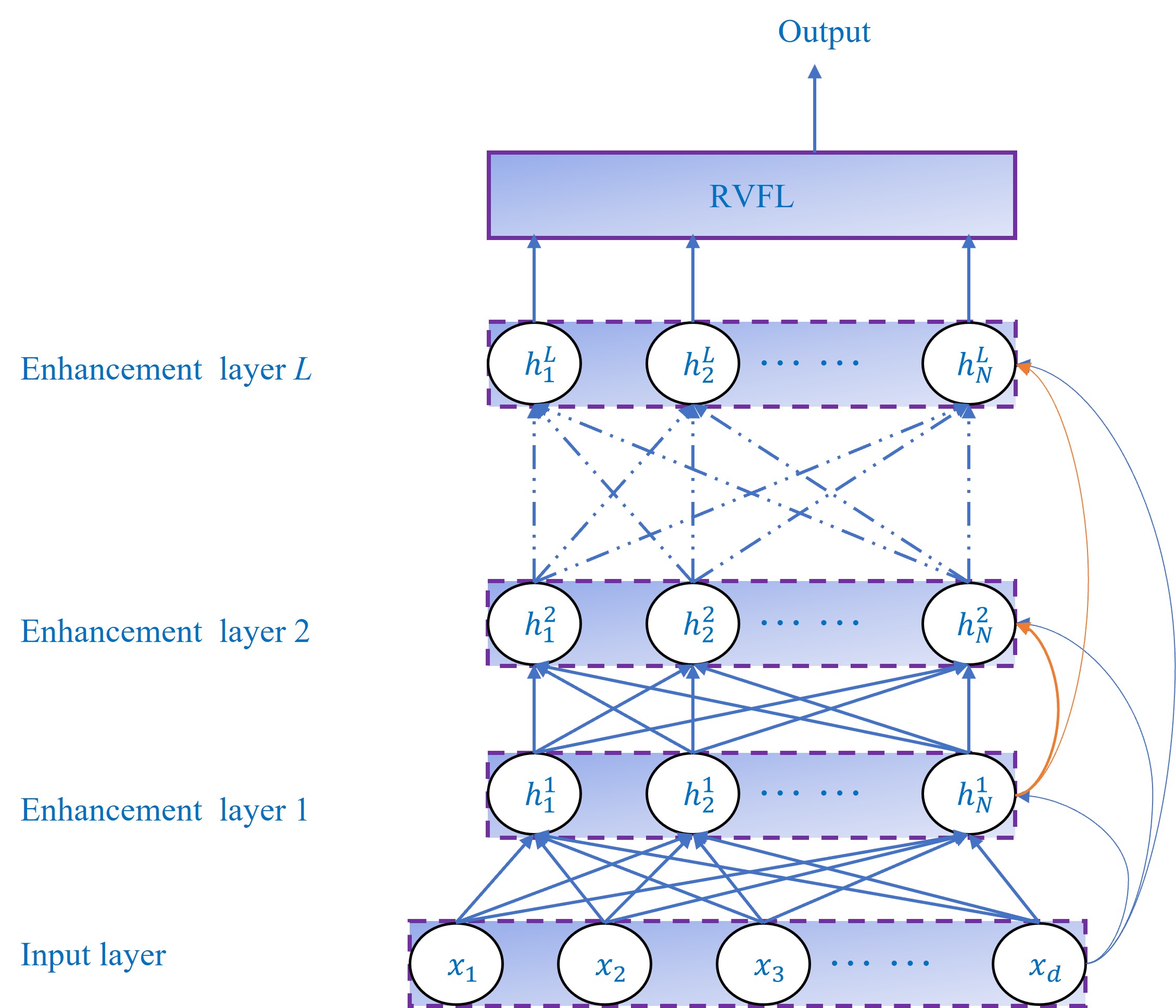}}
	\label{fig:sdRVFL(dense)}
\end{figure*}

\begin{figure*}
  \ContinuedFloat 
\subcaptionbox{edRVFL \cite{shi2021random}}{
	\includegraphics[width=.85\textwidth]{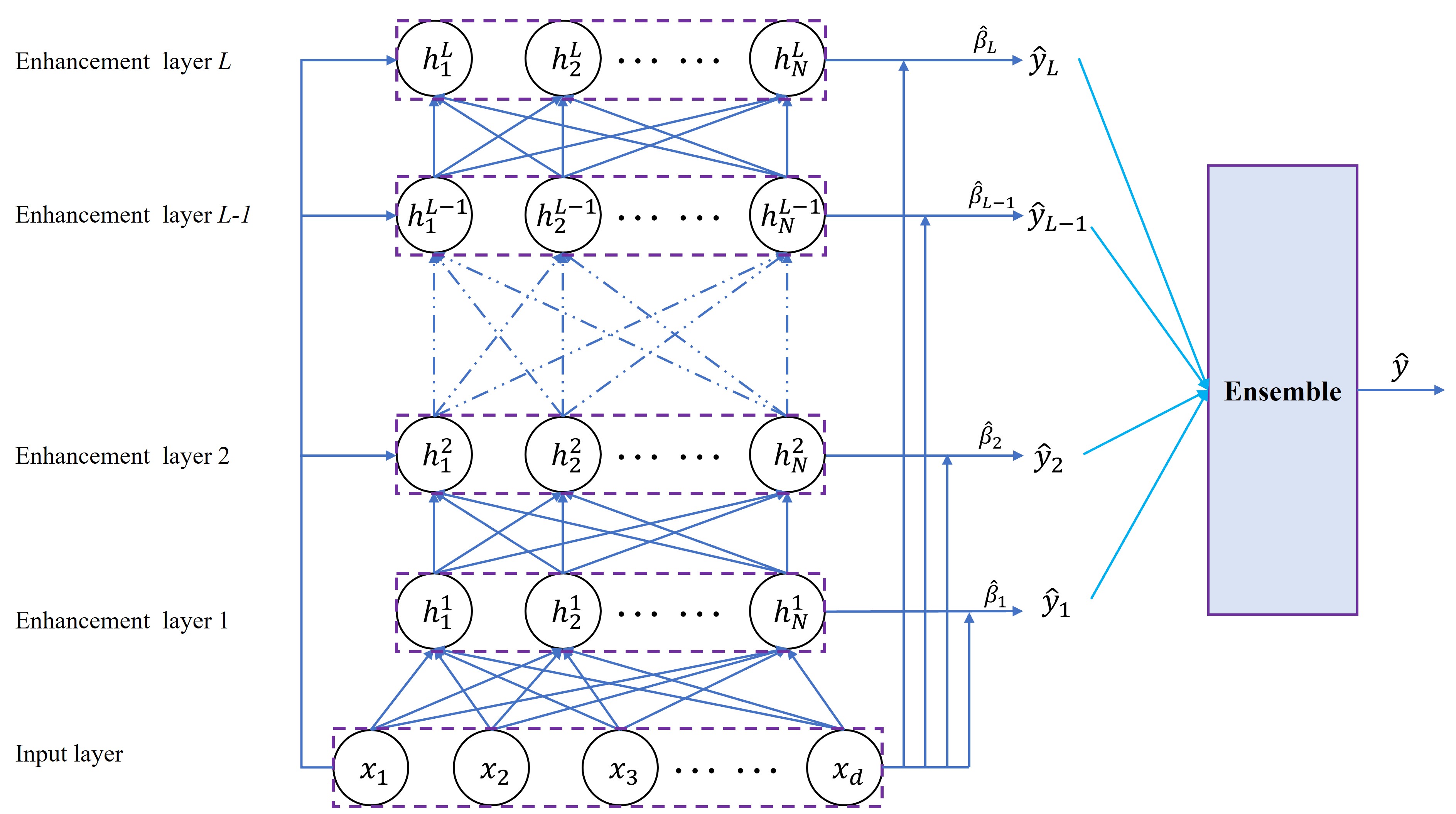}}
	\caption{Different types of architectures of deep RVFL model.}
		\label{fig:edRVFL}
\end{figure*}

\subsection{Stacked deep RVFL}
The stacked deep RVFL utilizes multiple enhancement layers to achieve multi-scale feature extraction. The consistent characteristic of the deep RVFL architectures is the multiple stacked enhancement layers. The main difference among them lies in how the direct links are connected. The sdRVFL is the most straightforward deep RVFL architecture, which stacks multiple hidden layers, and the direct link only connects the input layer and output layer \cite{henriquez2018twitter,katuwal2019stacked,zhang2016visual}. The convolutional deep RVFL also establishes the direct connection in this fashion, but the enhancement layers are convolution layers with random weights \cite{zhang2016visual,dai2021sar}. However, the direct links are not equipped with hidden layers, which weakens the unsupervised feature extraction. The direct links are densely connected to the hidden layers to guide the random features' generation, and the sdRVFL(dense) is proposed \cite{katuwal2019stacked}. The above architectures only utilize the last hidden layer's features for decision, which may lose valuable information from the intermediate layers. A dRVFL is proposed to take advantage of the rich information from all hidden layers in \cite{shi2021random}. A reservoir layer is utilized to extract features for the following deep RVFL \citep{hu2022deep}.

\subsection{Ensemble deep RVFL}
The stacked deep RVFL does not fully use the features from intermediate layers. However, the dRVFL's utilization of all features requires an inversion of a super large matrix \cite{shi2021random}. Therefore, the edRVFL is proposed to achieve a trade-off between computational efficiency and features utilization \cite{shi2021random,gao2021random}. The structure of edRVFL is shown in Figure \ref{fig:edRVFL}. In edRVFL, the direct link connects each enhancement layer with the input layer to guide the random features' generation. An individual output layer with a direct link to the input layer is built for each enhancement layer. Such design splits the inversion of a large matrix into multiple mini-matrix inversion and takes advantage of all features \cite{shi2021random}. After training all the output layers, an ensemble block generates the final output. The majority voting mechanism is adopted for classification tasks \cite{CHENG2021107826,shi2021random} and mean/median operation is applied for forecasting tasks \cite{gao2021random}. Recently, a comparative study shows that edRVFL outperforms ensemble deep ELM on human joint angles prediction \citep{yang2022deep}. In addition to using all hidden layers for decision making, \citet{yu2022selective} proposes to utilize a genetic algorithm for selection. Since the hidden neurons are randomized without optimization, there may be inferior neurons which hampers the generation of high-quality features in deeper layers. Therefore, \citet{shi2022weighting} prune the inferior neurons before generating the next layer's neurons. In addition, a weighting scheme, which assigns different weights to the training samples in each hidden layer, is proposed to improve the performance. The wrongly classified samples are assigned larger weights in the next layer to increase the diversity and accuracy simultaneously. The norm features from an edRVFL are concatenated with the privileged
information from another edRVFL with different activation \citep{ganaie2022ensemble}. This concatenation is fed into another deep RVFL for classifications. Recently, a strategy of random skip connections is proposed to enhance the representation ability of the edRVFL in \cite{hu2022edrvflR}. Instead of using all output layers, an edRVFL  with a selective  ensemble method is designed and succeeds in landslide displacement prediction \citep{yu2022selective}. Following the principle of determining important output layers, an edRVFL with dynamic ensemble based on online performance is proposed for time series forecasting \citep{du2022time}. An approach for the automatic design of ensemble deep randomized neural networks is proposed in \citep{hu2022automated}.

In addition to the supervised learning based on edRVFL, \citet{hu2022representation} propose a clustering algorithm based on edRVFL's features. The unsupervised learning is achieved by the manifold
regularization. Then, the $k$ means is developed based on the edRVFL features. The consensus clustering method is related to the ensemble block of the edRVFL. Recently, a novel edRVFL for semi-supervised tasks is proposed by \citet{shi2022jointly}. The proposed jointly optimized semi-supervised edRVFL (JOSedRVFL) minimizes the loss function consisting of three components, the error term, $L_2$ norm regularization and manifold regularization. The $L_2$ norm regularization aims at reducing the model's complexity, and the manifold regularization ensures the conditional probabilities of similar samples are close. In \cite{malik2022GE}, geometrical informations under the graph embedded framework are employed  while calculating the output parameters of each hidden layer (base model) and therefore, has better generalization performance than edRVFl model.
\begin{figure*}[htbp]
	\centering 
	\includegraphics[width=.8\textwidth]{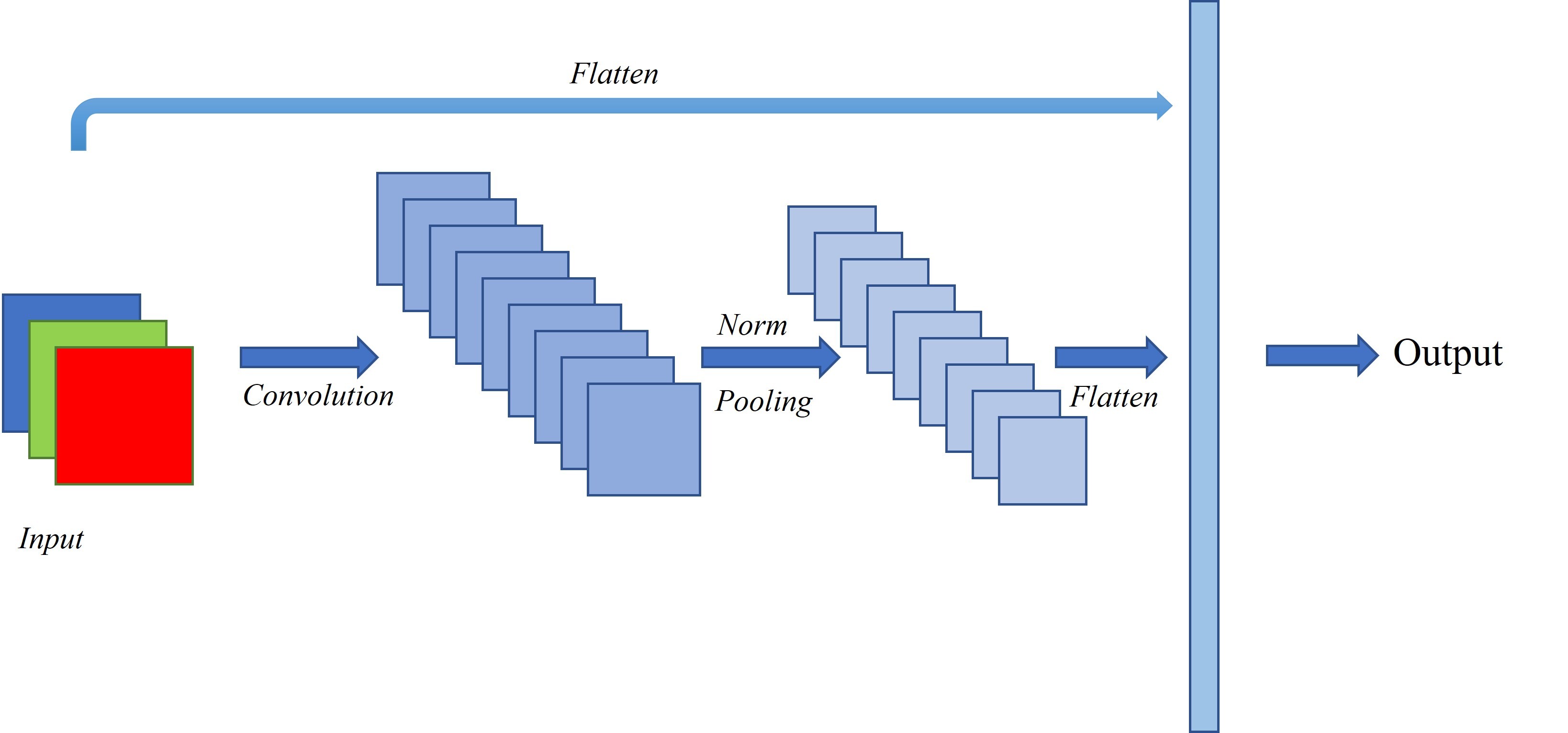}
	\caption{The architecture of deep convolutional RVFL.}
		\label{fig:ConvRVFL}
\end{figure*}
\subsection{Hybrid deep RVFL}
Unlike the above deep RVFLs, the hybrid ones utilize other advanced feature extractions techniques, like DL, to generate the input of the decision block. The decision block can be any RVFL's variants, including shallow and deep architectures. For instance, features extracted from a pre-trained ResNet-50 are fed into RVFL whose activation is s-fuzzy membership function \cite{sharma2021faf}. In \cite{CHENG2021107826}, the ResNet extracts features from time series data, and these features are fed into multiple edRVFL. The convolutional sparse coding deep network extracts features and feeds them into a stacked deep RVFL \cite{maeda2019convolutional}. In \cite{lu2021cerebral}, the base model in the ensemble learning framework is a hybrid deep RVFL whose DL performs feature extraction, and RVFL makes decisions.
\section{Hyper-parameters optimization and experimental setup}
 \label{Sec:Hyper-parameters optimization and experimental setup}
 The RVFL's performance heavily depends on hyper-parameters optimization. This section first separately summarizes the hyper-parameters optimization of single-layer and deep RVFLs. Finally, experimental setup, including data partitioning, evaluation metrics, and statistical tests, are presented.
 \subsection{Hyper-parameters optimization for single-layer RVFL}
 For the canonical RVFL, these hyper-parameters include input scaling, number of hidden nodes, activation functions, regularization strength, and distribution of random weights. Most literature utilizes a comprehensive grid search to tune these parameters. Grid search is straightforward to implement and succeeds on many tasks, such as classification \cite{chauhan2021randomized,zhang2016comprehensive}, forecasting \cite{gao2021walk,gao2021random}. A comprehensive evaluation of RVFL for classification is conducted by \citet{zhang2016comprehensive}. A grid search based on 4-fold cross-validation is utilized to select the number of hidden nodes and regularization strength for the RVFLs with different configurations. This study achieves some significant findings. First, the results demonstrate the superiority of the direct links. Second, the output layer's bias must be tuned based on the specific task. Third, the \textit{hadlim} and \textit{sign} activation functions usually degenerate the accuracy. Finally, tuning the scaling of randomization weights (input scaling) also increases the performance. \citet{gao2021walk} divides the time series into training, validation, and test set in chronological order. Then, a grid search is conducted to select the hyper-parameters, number of hidden nodes, and regularization strength, according to the forecasting errors on the validation set. Choosing the optimal activation function and number of hidden neurons is also an challenging job so some researchers adopt incremental learning techniques \cite{dai2022incremental} and kernel trick \cite{chakravorti2020non} to avoid  these issues. Table \ref{Table:activation function} shows the list of activation functions used in the literature. 
 
 However, exhaustive grid search has several drawbacks. Some literature implements evolutionary algorithms for hyper-parameters optimization, such as levy flight based PSO \cite{dash2021short} and chimp optimization \cite{zayed2021predicting}. The evolutionary algorithms encode the hyper-parameters into an individual, obtaining the optimal configuration after many generations. Each generation selects the individuals whose performance is outstanding. Therefore, generation after generation, the suitable configuration survives. 
 
 Based on the above descriptions, there are two main branches of hyper-parameters tuning of single-layer RVFL, the grid search, and evolutionary optimizations. The performance of grid searches heavily relies on the researchers' experience because they have to define the hyper-parameters for selection manually. A detailed discussion and analysis of single-layer RVFL's hyper-parameters are given in \cite{zhang2016comprehensive}. For evolutionary optimization of the RVFL, the researchers can define a large search region to allow the evolutionary algorithm to explore the best configuration. However, the evolutionary algorithms are time-consuming because each generation trains multiple RVFLs.
 
 The following suggestions about tuning single-layer RVFLs are provided for researchers and practitioners. First, large hidden dimensions are preferred for large datasets with huge input dimensions. Second, tuning the input scaling parameter may boost the performance on specific tasks, although the most common practice is to set the scaling factor to one. Third, it is advisable to increase the hidden nodes when tuning the input scaling worsens the performance. Fourth, the regularization strength of ridge regression plays a critical role in improving performance.
 
 \subsection{Hyper-parameters optimization for deep RVFL}
 As the RVFL becomes deep, the hyper-parameters that wait to be optimized grow exponentially. Whether each layer needs a different set of hyper-parameters is still an open problem. \citet{katuwal2019stacked} utilizes a grid search to optimize the number of hidden nodes and regularization, but the number of layers is fixed as three. 
 
 Some hyper-parameters tuning strategy is proposed for the deep RVFLs. For instance, a two-stage tuning strategy is proposed to obtain the best configurations of edRVFL for classification \cite{shi2021random}. The optimal number of hidden nodes and regularization parameters are selected for a two-layer network in the first stage. Optimizing the hyper-parameters for a two-layer network saves computational time and also considers the effects of deep representations. Then, the second stage fine-tunes the hyper-parameters within the neighborhood of the obtained number of hidden nodes and regularization parameter from the first stage optimization. Similar to the two-stage tuning, a three-stage method is proposed by \citet{CHENG2021107826} for time series classification. In \cite{CHENG2021107826}, the authors tunes the number of hidden nodes and regularization parameters for the first and second hidden layer in the first and second tuning stages. Then, the third stage imposes a random deviation on the optimal hyper-parameter obtained from the previous steps. The random variations can enhance the generalization ability and reduce the computational burden of the hyper-parameters tuning. \citet{gao2021random} proposes a layer-wise grid search algorithm to determine the deep RVFL's hyper-parameters layer by layer. The cross-validation is conducted to obtain the best hyper-parameters. Once the hyper-parameters tuning is finished, the hyper-parameters and hidden states for this layer are fixed. The cross-validation is applied to the next layer. This process is repeatedly until the hyper-parameters for the last layer are determined.The layer-wise tuning offers each layer an opportunity to own different configurations. The layer-wise tuning method is also utilized in \cite{gao2022inpatient}.
 
 When the neural networks become deep, the hyper-parameters tuning becomes more challenging, because there are more hyper-parameters and the computation burden increases. For the deep RVFLs, the tuning of hyper-parameters must take into account both computational efficiency and performance. Hence, to achieve these two goals, stage-based \cite{shi2021random,CHENG2021107826} and layer-wise tuning algorithms are proposed \cite{gao2022inpatient,gao2021random}. In summary, the stage-based tuning divides the tuning process into stages to save computation time and ensure efficient hyper-parameter exploration. The first stage of stage-based tuning searches hyper-parameters within a coarse region for saving computation time and exploring a large space. The following stages are the fine tuning to further improve the accuracy. The layer-wise tuning considers each hidden layer and the corresponding output layer as an independent model. Hence, each layer's tuning only works on a single-layer RVFL whose computation is much faster than that of deep architectures. Furthermore, such tuning ensures that each output layer performs exceptionally well. Finally, the layer-wise tuning can offer different layers with different hyper-parameters to increase the diversity. 
 
 The preceding discussion offers some insights into tuning hyper-parameters for the deep RVFLs. First, tuning for shallow RVFLs or each layer can save computational time. Second, searching within a coarse region is an efficient way to explore distinctive hyper-parameters. Third, fine-tuning based on the selected hyper-parameters from wide ranges can further improves the performance. Fourth, assigning different hyper-parameters to each layer contributes to the diversification of the edRVFL.

 \subsection{Experimental setup}
 This section presents and summarizes the experimental procedures, including data partitioning, normalization, evaluation metrics, and statistical tests. For classification problems, the researchers usually adopt the $k$-fold cross-validation for hyper-parameters tuning and evaluate the models on the remaining test set \cite{zhang2016comprehensive}. If the partition schemes of the dataset are available, the researchers must follow the same partitioning for fair comparisons, such as annealing and audiology-std dataset \cite{zhang2016comprehensive}. 
With a given experimental setup, there is a need to benchmark the models for fair comparison. There have been multiple attempts in the literature to benchmark the performance of the models on a given experimental setup \citet{fernandez2014we}. Recently, self normalizing networks \cite{klambauer2017self} released the publicly available data partitions for reproducibility and the benchmarking of the models. Following self normalizing networks [self normalizing ref], several studies like \cite{shi2021random} followed this setup for fair evaluation of the models.  Recently, \citet{del2022randomization} presented a through survey based on randomization based machine learning models with renewable energy prediction problems and compared them. There is still a gap for the benchmarking of the models like lack of evaluation of the models across different hyperparameters, their range and so on. Thus, benchmarking of the models to ensure the progress of the literature, reproduciblity of the results and fair comparison is needed in randomization based models.

 The test set is always located at the end for time series datasets, and the remaining observations are utilized for training and hyper-parameters tuning. There are two approaches to split the training and validation set. The first approach is the same as the cross-validation for regression and forecasting. Some researchers formulate the observations into input patterns and response values. Then, a $k$-fold cross-validation is conducted to tune the hyper-parameters \cite{qiu2016electricity}. The second approach splits the training and validation set in chronological order \cite{gao2021walk}.

  All models have limitations, and therefore evaluation and comparison of machine learning models depend on the specific dataset. A fruitful set of evaluation metrics is utilized to evaluate the RVFL's performance. For classification, the classification accuracy is always the first choice \cite{shi2021random}. Table \ref{tab:Summary of forecasting errors} summarizes the forecasting errors utilized in the literature, where, $x_{j}$ and $\hat{x}_{j}$ represent the raw observation and its forecast, $L$ and $T$ represent the size of training and test set, respectively. MAE and RMSE can be utilized when the time series are of the same scale, although RMSE is more sensitive to outliers. MAPE is a popular percentage error with high interpretibility. Finally, MASE is a scaled metric and can be utilized for comparisons on different time series. In addition to the forecasting errors, the direction statistics are utilized for comparison in some literature \cite{tang2018non,tang2018randomized}. The definition of direction statistics (\textit{Dstat}) is 
  \begin{equation}
  Dstat=\frac{1}{L}\sum_{1}^{L}a_{i}\times 100\%,
  \end{equation}
  where $a_{i}=1~ \text{if}~ (\hat{x}_{j}-x_{j-1})(x_{j}-x_{j-1})>0$, or otherwise $a_{i}=0$.
  \begin{table}[htbp]
  \centering
    \caption{Forecasting errors in the literature about RVFL.}
    \label{tab:Summary of forecasting errors}
    \small
    \begin{tabular}{lll}
    \hline
Metric&Formula \\
\hline
Mean absolute error (MAE)&$\frac{1}{L}\sum_{1}^{L}|\hat{x}_{j}-x_{j}|$ \\
Mean absolute scaled error (MASE)&$\frac{1}{L}\sum_{1}^{L}\frac{|\hat{x}_{j}-x_{j}|}{\frac{1}{T-1}\sum_{t=2}^{T}|x_{t}-x_{t-1}|}$ \\
Mean squared error (MSE)&$\frac{1}{L}\sum_{1}^{L}(\hat{x}_{j}-x_{j})^{2}$\\
Root mean squared error (RMSE)&$\sqrt{\frac{1}{L}\sum_{1}^{L}(\hat{x}_{j}-x_{j})^{2}}$\\
Mean absolute percentage error (MAPE)&$\frac{1}{L}\sum_{1}^{L}\left| \frac{\hat{x}_{j}-x_{j}}{x_{j}} \right|.$\\
\hline
\end{tabular}
\end{table}
  
  In addition to the above evaluation metrics, the literature also utilizes statistical tests to compare the different models' performance on various datasets. In general, these tests can be classified into two groups, the group-wise and pair-wise tests. Group-wise tests can determine the overall ranking of the models on all the datasets and group them based on the statistical distance. The literature about RVFL utilizes the Nemenyi test to compare the models in a group-wise fashion \cite{gao2021walk,qiu2016electricity}. The pair-wise tests assist in comparing the models in a pair-wise fashion, which is straightforward to show the better model. For instance, the Wilcoxon test is utilized to ascertain how many algorithms edRVFL significantly outperforms \cite{shi2021random}. There are several others statistical tests such as Friedman test, sign-test and so on, to evaluate the performance of machine learning models. We refer the reader to \cite{demvsar2006statistical,fernandez2014we,carrasco2020recent} for more detailed information about the application of these tests to machine learning models.


\section{Time series forecasting and other applications}
\label{Sec:Other applications of RVFL model}
Time series forecasting refers to establishing the model using historical observations, and this model is utilized to make extrapolations for future steps. Accurate and reliable forecasts help the stakeholders and decision-makers plan, organize, maintain and develop the system in advance in a data-driven fashion. 
Table \ref{tab:Summary of RVFL for forecasting} summarizes the representative literature about forecasting by RVFL and its variants.
RVFL and the improved versions have demonstrated their outstanding performance on various forecasting tasks from different domains, such as electricity load \cite{gao2021walk,qiu2016electricity}, solar power \cite{aggarwal2018short}, wind power \cite{qiu2017short}, financial time series \cite{qiu2018ensemblecrude} and other data \cite{hazarika2020modelling}. Table \ref{tab:Summary of RVFL for forecasting} shows that signal decomposition algorithms are popular for feature extraction on forecasting tasks. The signal decomposition can splits the time series into multiple sub-series with different frequencies. Then the RVFL works on these sub-series for forecasting. This section presents the details of all the literature about RVFL-based forecasting.

\begin{table}[htbp]
    \centering
    \caption{Summary of RVFLs for forecasting}
    \label{tab:Summary of RVFL for forecasting}
    \resizebox{\textwidth}{!}{
    \begin{tabular}{llllll}
    \hline
         Year&Literature&Feature extraction&Learning category&Hyper-parameter optimization&Field  \\
         \hline
         2022&\citet{gao2022inpatient}&-&edRVFL&Layer-wise grid search&Inpatient discharges\\
         \hline
         2021&\citet{zayed2021predicting}&-&Kernelized RVFL&Chimp Optimization Algorithm&Solar power  \\
         \hline
         2021&\citet{dash2021short}&Signal decomposition&Expanded RVFL&Particle swarm optimization&Solar power  \\
         \hline
         2021&\citet{majumder2021real}&-&Online sequential kernel RVFL&-&Solar power  \\
         \hline
         2021&\citet{gao2021walk}&Signal decomposition&Single model&Grid search&Electricity load \\
         \hline
         2021&\citet{hu2021short}&-&Ensemble RVFL&Evolutionary optimization&Electricity load  \\
         \hline
         2021&\citet{manibardo2021random}&Deep architecture&RVFL, deep RVFL and edRVFL&Bayesian optimization& Road traffic\\
         \hline
         2020&\citet{cheng2020new}&Signal decomposition&Decomposition-based ensemble learning&-&Wind speed  \\
         \hline
         2018/2020&\citet{qiu2018ensemblecrude,tang2018non,zhang2020novel}&Signal decomposition&Decomposition-based ensemble learning&Grid search&Crude oil price \\
         \hline
         2020&\citet{wu2020daily}&Signal decomposition&Decomposition-based ensemble learning&Sine cosine algorithm&Crude oil price \\
         \hline
         2020&\citet{wu2020hybrid}&Signal decomposition&Decomposition-based ensemble learning&whale optimization algorithm&Financial time series \\
         \hline
         2020&\citet{zhang2020enhancing}&Stacked auto-encoder&Incremental RVFL&Grid search&FCCU end-point quality \\
            \hline
               2020&\citet{hazarika2020modelling}&Signal decomposition&Single RVFL&Grid search& COVID-19 cases\\
               \hline
         2019&\citet{bisoi2019modes}&Signal decomposition&Single RVFL&Grid search&Crude oil price \\
         
         \hline
         2019&\citet{KUSHWAHA2019124}&Signal decomposition&Decomposition-based ensemble learning&Grid search&Solar power  \\
         \hline
         2019&\citet{majumder2019short}&-&Kernelized RVFL&Water cycle algorithm&Solar power  \\
         \hline

          2018&\citet{moudiki2018multiple}&-&RVFL with different regularizations&Grid search&Financial time series \\
          \hline
           2018&\citet{LIAN20181}&-&Ensemble RVFL&Grid search&Landslide displacement \\
           \hline
           2018&\citet{li2018travel}&Signal decomposition&Decomposition-based ensemble learning&Grid search&Travel time \\
           \hline
            2018&\citet{lu2018ensemble}&-&Ensemble RVFL&Grid search&Production rate \\
            \hline
            2017/2018&\citet{qiu2017short,nhabangue2018wind}&Signal decomposition&Decomposition-based ensemble learning&Grid search&Wind power  \\
         
         \hline
         2016/2018&\citet{qiu2016electricity,qiu2018ensembleload}&Signal decomposition&Decomposition-based ensemble learning&Grid search&Electricity load\\
         \hline
            2016&\citet{zhang2016multivariable}&-&Incremental RVFL&Grid search&Molten iron quality \\
            \hline
            2015&\citet{zhou2015multivariable}&PCA&Online sequential RVFL&Grid search&Molten iron quality \\
            
    \hline
    \end{tabular}}
\end{table}

\subsection{Electricity load}
Electricity load forecasting is crucial for the development, maintenance, and planning of power systems. Among the abundant forecasting methods, RVFL demonstrates its success by many researchers. For instance, a quantile scaling method is proposed to re-distribute the randomly weighted inputs of RVFL to avoid the saturation effects and suppress the outliers in \cite{ren2016random}. Signal decomposition techniques are utilized to remedy the unsupervised features of RVFL in \cite{gao2021walk,qiu2018ensembleload,qiu2016electricity}. For instance, EMD decomposes the load data into several modes, and then each mode is predicted by an RVFL. Finally, the summation is conducted to combine the predictions for each mode \cite{qiu2016electricity}. In \cite{qiu2018ensembleload}, a two-stage decomposition method is proposed to decompose the load data, then each load is predicted by an individual RVFL, and finally, all forecasts are aggregated by another RVFL with explanatory variables. Different from the above ensemble methods, a single RVFL is built on all the components generated by EWT in \cite{gao2021walk}. The same decomposition scheme is combined with edRVFL for short-term load forecasting in \cite{gao2021random} and the results demonstrate edRVFL's superiority over a single RVFL. Moreover, a multi-modal evolutionary algorithm is utilized to optimize the enhancement weights, bias, and combination weights of the ensemble RVFL for short-term load forecasting \cite{hu2021short}.
\subsection{Solar power}
With renewable energy development, solar power forecasting is an emerging area. In \cite{aggarwal2018short}, the authors compare RVFL with SLFN and RWSLFN, and the results demonstrate the superiority of the direct link. Signal decomposition methods also work for solar time series. For instance, in  \cite{KUSHWAHA2019124}, maximum overlap DWT decomposes the power data into sub-series, and an individual RVFL predicts each series. Finally, aggregation of all forecasts is the forecast for solar power. Moreover, some researchers utilize meta-heuristics algorithms to optimize RVFL's parameters automatically. For instance, in \cite{majumder2019short}, the multi-kernel RVFL whose kernel parameters are optimized by water cycle algorithm is proposed to forecast short-term solar power. In \cite{zayed2021predicting}, Chimp Optimization Algorithm (CHOA) is utilized to determine RVFL's hyper-parameters for predicting output power and the monthly power production of a solar dish/Stirling power plant. Some researchers integrate signal decomposition, evolutionary optimization, and RVFL together to boost forecasting accuracy. For example, in \cite{dash2021short}, the EWT is utilized to decompose the time series, and the residue is discarded. The remaining sub-series are expanded using trigonometric activation in the direct link, and the enhancement states are a linear combination of two activation functions. Finally, the RVFL is trained with a novel robust objective function by minimizing the variance of training data. Moreover, the added activation functions' hyper-parameters are also optimized by PSO. The new cost function also shows its improvement on RVFL in terms of forecasting accuracy. In \cite{majumder2021real}, an Online Sequential Kernel-based Robust RVFL is trained based on Hampel's cost function to forecast solar and wind power.
\subsection{Wind power}
A comparison of RVFL and other machine learning models on wind speed forecasting is conducted in \cite{gupta2021short}. Some literature about wind power forecasting combines signal decomposition techniques with RVFL \cite{nhabangue2018wind,qiu2017short}. For instance, CEEMD is applied to decompose the raw data into modes, and a kernel ridge regression predicts each mode. Finally, instead of using a simple summation, the RVFL is trained to combine the forecasts of all methods for wind power ramp prediction in \cite{qiu2017short}. In \cite{nhabangue2018wind}, Chebyshev expansion is utilized as functional nodes to reduce the number of activation nodes. Then it is combined with EMD for wind speed forecasting. In addition, Hampel's cost function is utilized for training an online sequential kernel-based robust RVFL to forecast solar, and wind power in \cite{majumder2021real}. In \citet{cheng2020new}, a multi-objective salp swarm optimizer is adopted to determine the weights which are used to combine the forecasts from four networks, including RVFL, for wind speed forecasting. 

\subsection{Financial time series}
The financial time series is different from the above data with strong cycles. The financial time series is much volatile, and it is difficult to extract features. Among all the RVFL-related financial time series forecasting literature, most focus on crude oil prices. Similar to the other kinds of time series forecasting literature, many researchers combine RVFL and signal decomposition algorithms for crude oil price forecasting \cite{zhang2020novel,qiu2018ensemblecrude,tang2018non}. EEMD decomposes crude oil price, and then different RVFLs are trained for each mode, including the residue. Finally, the summation of all RVFL's outputs is the forecast in \cite{tang2018non}. In \cite{qiu2018ensemblecrude}, CEEMD decomposes the raw data into modes, and an individual RVFL is established on each mode. Finally, forecasts of all modes are aggregated using an incremental RVFL. The same decomposition-based structure is utilized. The difference is that improved CEEMD with adaptive noise acts as the decomposition in \cite{wu2020hybrid,wu2020daily} and sine cosine algorithm optimizes all the parameters in\cite{wu2020daily}. In \cite{zhang2020novel}, bivariate EMD is utilized to decompose the original time series into sub-series, and an individual RVFL predicts each series. Finally, aggregate the forecasts via summation.
The modes generated from VMD are fed into RVFL, and the experimental results demonstrate the superiority of VMD over EMD in \cite{bisoi2019modes}. Besides the decomposition-based RVFL, a novel ensemble RVFL with five diversity strategies is proposed for crude oil price forecasting in \cite{yu2021investigation}.
   
 Besides the literature about crude oil price forecasting, RVFL also succeeds on other financial time series. For example, in \cite{moudiki2018multiple}, different regularization parameters are imposed to the output weights of the direct link and enhancement nodes to forecast discount rates. In \cite{wu2020hybrid}, the improved CEEMD with adaptive noise decomposes the data into sub-series, and each sub-series is predicted by an individual RVFL whose parameters are optimized by a whale optimization algorithm. Finally, the output is the summation of all forecasts. In \cite{li2020forecasting}, multilingual search engine data is utilized to derive the input for RVFL to forecast crude oil price. 
\subsection{Other Applications}
Besides the above popular areas with a large need for forecasting, RVFL and its improved versions have also succeeded in various areas, such as temperature \cite{zhang2019robust}, landslide displacement \cite{LIAN20181}, COVID-19 cases \cite{hazarika2020modelling}, travel time \cite{li2018travel}, molten iron quality \cite{zhang2016multivariable,zhou2015multivariable}, energy consumption \cite{sun2021privileged}, signal-to-noise ratio \cite{xue2020rvfl}, algae missing values \cite{hussein2019new}, temperature in subway station \cite{zhang2018forecast}, inpatient discharges \cite{gao2022inpatient} and so on \cite{manibardo2021random}. 

Among these literature, many utilize different heuristic algorithms to optimize the hyper-parameters \cite{abd2021new,abd2020utilization,ESSA2020322,sharshir2020enhancing}, weights \cite{LIAN2020286} or select the input features \cite{hussein2019new}. For instance, firefly algorithm is utilized to select RVFL's hyper-parameters, number of enhancement nodes, bias, direct link, distribution and activation function, for thermal performance prediction \cite{hazarika2020modelling}.
In \cite{chen1998incremental}, an incremental method to adjust RVFL's structure is proposed for time series prediction, where the network increases its enhancement nodes when the performance degrades.
 For instance, in \cite{hussein2019new}, a moth search algorithm is utilized to select input features for RVFL to predict missing values of algae. In \cite{abd2020utilization}, Marine Predators Algorithm is utilized to optimize RVFL's hyper-parameters for tensile behavior prediction. In \cite{ESSA2020322}, an artificial ecosystem-based optimization algorithm is utilized to optimize RVFL's hyper-parameters for forecasting power consumption and water productivity of seawater. In \cite{abd2021new}, mayfly-based optimization is utilized to optimize RVFL's hyper-parameters to forecast the performance of Photovoltaic/Thermal Collector. In \cite{LIAN2020286}, the RVFL trained by PSOGA is utilized to generate prediction intervals for landslide displacement. The RVFL is first pre-trained using reconstructed intervals, and then the PSOGA trains the RVFL with transferred weights based on original data. 
 
 Besides the RVFL-based on meta-heuristics, some literature also focuses on ensemble RVFL \cite{li2018travel,sun2021privileged,lu2018ensemble,LIAN20181}. For instance, GA is utilized to assign ensemble weights for each RVFL trained with bootstrap samples, and the RVFLs whose weights are higher than the threshold are selected to construct prediction intervals for landslide displacement \cite{LIAN20181}. In \cite{li2018travel}, EEMD is utilized to decompose the travel time into modes, and a different RVFL predicts each mode. Finally, each mode's forecasts are combined with linear addition. In \cite{zhang2019robust}, ensemble RVFL is trained based on AdaBoost after selection features via MRMR. Each RVFL is trained using iteratively reweighted least squares for temperature forecasting. In \cite{sun2021privileged}, EEMD decomposes time series into sub-series, and the features with high correlation with target variables are used for the corresponding RVFL to forecast building energy consumption. In \cite{lu2018ensemble}, Negative Correlation Learning is utilized for training ensemble RVFL networks for production rate forecasting. \citet{manibardo2021random} applies RVFL, deep RVFL, and edRVFL to the road traffic dataset, and the hyper-parameters are determined by Bayesian optimization. \citet{manibardo2021random} claims that the direct link is the fundamental reason for RVFL and its variants' superiority over ELM-based models. 
 
 Incremental (online) RVFL also succeeds on various time series \cite{dash2018indian,zhang2016multivariable,zhou2015multivariable,zhang2020enhancing,qu2018rapid}. Incremental RVFL updates its structure or weights when new observations are available. For example, in \cite{zhang2016multivariable}, the incremental RVFL adds a new node and updates its weight incrementally until the performance degrades for the prediction of molten iron quality. In \cite{zhou2015multivariable}, the online sequential RVFL is trained using the principal components and its estimation of the previous steps to predict molten iron quality, too. In \cite{dash2018indian}, RVFL and online sequential RVFL are compared on rainfall prediction, and the results demonstrate OS-RVFL's superiority for rainfall forecasting. In \cite{qu2018rapid}, an online RVFL-based on sliding-window is trained to temperature forecasting. In \cite{zhang2020enhancing}, a stacked auto-encoder is trained in offline fashion first, and then an incremental RVFL is established based on the SAE's output when a concept drift is detected. 
   
 In addition, different novel RVFLs are proposed for other time series. In \cite{hazarika2020modelling}, the level one sub-series generated from discrete wavelet transformation are fed into RVFL for COVID-19 cases forecasting. In \cite{ZHOU20191}, Schmidt orthogonalization is utilized to orthogonalize the output vectors, and the hidden nodes are pruned according to the output weights to predict product quality. 
 
 Although RVFL and its variants succeed in various time series forecasting tasks, the research on spatial-temporal time series is not mature. There is only one RVFL-related paper touching this problem to the authors' best knowledge. In \cite{xu2017kernel}, a kernel RVFL is established to predict the temporal dynamics decomposed via Karhunen–Loève method from the spatial-temporal process.

\section{Comparison with other state of the art machine learning techniques}
\label{Sec:comparing with SORT}
An insightful discussion about the comparison between the RVFLs and other state of the art machine learning techniques significantly contributes to the literature. This section mainly discusses about the pros and cons of the RVFLs and other machine learning models. 

The early work with randomization techniques can be traced with perceptron and standard feed-forward neural network (SLFN). In perceptron, the weights between sensor units and response units can be generated randomly whereas the rest weights from the associator units and the response units are calculated via reinforcement learning  \cite{rosenblatt1958perceptron,rosenblatt1961principles}. SLFN \cite{schmidt1992feed} also uses randomization technique but there is no direct links in this network. Jacobian neural Network (JNN) \cite{elisseeff1999jnn} is a polynomial time randomized algorithm which give optimal network with probability one.  Moreover, the paper \cite{igelnik1995stochastic} has some theoretical justification for RVFL and other neural networks.
Back-propagation based trained ANNs are sensitive to learning rate setting, slow convergence and trapped into local minima \cite{jacobs1988increased,gori1992problem,magoulas1999improving}. On the other hand, RVFL resolves these issues by generating the weights randomly from input layer to hidden layer and the rest weights (hidden layer to output layer) are calculated via closed form solution. The direct links in RVFL plays an important role in both classification and regression problem \cite{zhang2016comprehensive,vukovic2018comprehensive}. These direct connections separates RVFL from other randomized networks such as radial basis function (RBF) \cite{broomhead1988radial} and extreme learning machine (ELM) \cite{huang2006extreme} and so on.
RVFL and its deep variants have shown superior performance than ELM, Hierarchical ELM (H-ELM) and multi-layer kernel ELM (ML-KELM) \cite{katuwal2019stacked}.
Support vector machine (SVM) has strong mathematical foundation and has shown state of the arts results \cite{wang2005support,cortes1995support,tanveer2019comprehensive}. However, RVFL with privileged information (RVFL+) and its kernel extension (KRVFL+) have shown superior performance than SVM and its variants such as gSMO-SVM+ and fast SVM+ \cite{zhang2020new}. From optimization perspective, RVFL+ has simpler constraints than SVM+, that results in closed form solution. Furthermore, the DRVFL shows superiority over SVM and random forest on Twitter sentiment datasets. The DRVFL with fuzzy activation outperforms EML and kernel-ELM on the ADNI dataset.

One popular state of the art deep learning method is the Resnet. The Resnet constructs a quite deep architecture with the help of residual links. It utilizes back-propagation algorithm to optimize the weights and bias, which takes much more time than training the RVFL networks. In addition, the literature has demonstrated the superiority of deep RVFL over the Resnet on tabular data classification \cite{shi2021random,shi2022weighting}. Besides tabular datasets, time series forecasting is also a valuable problem. For forecasting, the long short-term memory (LSTM) and temporal convolutional network (TCN) are two common state-of-the-art methods. Comparing with edRVFLs, the training is much slower. However, many literature show that the advanced RVFLs outperform the LSTM and TCN \cite{gao2021random}. Furthermore, the deep stacked RVFL method outperforms stacked denoising auto-encoder on two benchmark MR brain datasets (MD-1 and MD-2). Some literature has demonstrated that RVFL-based models outperform the BPNN \cite{bisoi2019modes}.

However, RVFL does not include CNN-type feature extraction layers for image or sequence data. The convolution filters aim at mining local patterns from different spaces. Then, multiple stacks of these filters assist in learning a global representation. Finally, the gradient descent algorithms help to learn these features in an end-to-end fashion. Although RVFL does not own the CNN's feature extraction layers, the features learned by CNNs can be utilized as input to the RVFL variants. In other words, the RVFL variants can be the decision module for the features from all kinds of gradient-based deep networks \cite{CHENG2021107826}.

\section{Conclusions and future directions}
\label{Sec:Conclusion}
Randomized neural networks (RNNs) have shown their strength among machine learning models. A special kind of RNN, RVFL model, has been emerged as a very successful model. The history of RVFL model can be traced in late 20th century. 
This review paper summarizes the developments of RVFLs from theoretical foundations to various applications. As per our knowledge, this is the first review paper focused for RVFL model. The RVFL is a feed-forward neural network with random features and direct links. The randomized features introduce non-linear representations of the input features, and the direct links reserves the linear pattern. The hidden layer's weights are randomly initialized and frozen during training, and only the output layer is trained with a closed-form approach. The randomized features render RVFLs at a fast computational speed. 

With the renaissance of deep learning, researchers have extend the shallow RVFL to deep architectures to enhance its representation ability. In the deep architectures of RVFL, the hidden neurons are also randomly initialized and frozen during the training step. Only the output layers are trained, which reduces the computational burden of back-propagation. The RVFLs with deep architectures have demonstrated their superiority over shallow ones on classification, regression, and forecasting, etc. Therefore, the authors claim that the hierarchical enhancement features offer a large modeling capacity and increase performance.    Different ensemble learning algorithms, such as bagging, boosting, and stacking, are shown to significantly boost the single RVFL's performance. In addition, the ensemble RVFLs based on signal decomposition demonstrate tremendous success on various forecasting tasks. The signal decomposition algorithms first dis-aggregate the complex sequential data into multiple components, which assists in the RVFL's representation ability. Then RVFL-based models are built on each component, and the ensemble of all forecasts is the output.  
The RVFL-based models have achieved significant success in various domains because of their fast computational speed, high accuracy, and powerful representation ability and these models also have achieved state-of-the-art performance in the time series forecasting domain on wind speed, solar energy, electricity load, etc. we hope that this paper offers treasure information about the RVFL model to the researchers.
we presented a thorough survey on the developments of the RVFL model in many aspects, i.e., shallow RVFL, ensemble algorithms based on the RVFL model, deep RVFL variants, etc. Also, we presented applications of the RVFL model that shows its applicability in the real world. The literature has demonstrated the superiority of the RVFL-based deep models over tabular datasets \cite{shi2022weighting}. While reviewing the papers in the literature, we found some potential research directions that the researchers in future should explore. 
\begin{itemize}
\item Weights initialization techniques (WITs) have significant impact over the performance of RVFL models. A few research \cite{zhang2016comprehensive,cao2017impact,
cao2019initial,tanveer2021ensemble} suggest that this topic needs to be discussed further with mathematical justifications. Moreover, several others strategies \cite{narkhede2022review} can be explored with RVFL model such as interval based initialization \cite{sodhi2014new}, variance scaling based initialization \cite{yang2021new}, data-driven initialization \cite{li2017initializing}, hybrid initialization \cite{mishkin2015all,aguirre2019improving}, cluster based initialization \cite{steiner2022cluster} and data statistics based initialization \cite{koturwar2017weight}.
\item Outliers or noisy samples influence the modeling capability of standard RVFL and hence, leads to poor performance. Kernalized RVFL models are robust but can't be employed for large scale dataset. Therefore, for large scale, different techniques such as random Fourier features \cite{rahimi2007random} can be used to handle the same \cite{mehrkanoon2018deep}. Moreover, RVFL with fuzzy neural networks  \cite{de2020fuzzy}, ensemble learning \cite{ren2016ensemble} or other advanced techniques can be employed to develop robust RVFL variants.
  \item Ensemble learning and deep learning are two individual growing fields \cite{ganaie2021ensemble}. Researchers have recently combined them to develop a more accurate and efficient model that can perform well on real-world data. The RVFL model has fast training speed and good generalization performance and has been employed successfully in various engineering domains. Therefore, this can be a hot topic for researchers to explore RVFL in these research directions. 
  \item In the literature \cite{katuwal2018enhancing, katuwal2018ensemble}, RVFL and decision tree have been employed together to develop a model with better performance. Recently, deep forest \cite{zhou2017deep,zhou2019deep} with better interpretability and less tunable parameters as compare to deep neural networks (DNNs) is a growing research field. Studying the RVFL model and decision tree with deep forest architecture can be a new research field.
  \item To increase the generalization performance of machine learning models, learning with global/local data consistency (topological properties of data) has shown its importance among the machine learning community. The RVFL model transforms the original features into randomized features in unsupervised manner. Hence, the randomization process might ruin the original feature space's topological properties and leads to an inefficient model. Therefore, works \cite{ganaieminimum,li2021discriminative} indicates that incorporating the idea of manifold learning into the RVFL model can develop more accurate models.
  \item  The standard RVFL handles balanced data effectively. The imbalance learning problem seriously deteriorates the performance of the RVFL model. In general, techniques addressing imbalance data are divided into two categories, i.e., data-level approach \cite{chawla2002smote} and algorithm approaches \cite{lopez2012analysis} can be used with RVFL model to classify imbalance data, effectively. Therefore, it is an opportunity for researchers to develop other techniques to explore this research direction.
  \item Developments in the RVFL model have been focused on supervised problems, i.e., classification and regression problems. In the real-world, unlabeled data consist of only a few labeled samples but many unlabeled samples. There are very less work with RVFL to handle semi-supervised problems. Therefore, it is desirable to develop variants of the RVFL model that can be employed for semi-supervised problems effectively.
    \item The model pool of ensemble RVFL mainly consists of learning algorithms. However, statistical methods can be included and improve the performance. For instance, statistical forecasting methods, such as ARIMA and exponential smoothing can be included in the model pool for forecasting tasks. Therefore, the models' diversity is increased significantly.
  \item Most deep RVFLs networks are designed based on the conventional feed-forward architecture. However, the deep learning community has proposed various advanced architectures, such as the Transformer \cite{vaswani2017attention} and graph convolution neural network, etc. Combining the advanced architectures and the idea of deep RVFL may maintain high performance and reduce training time simultaneously.
  \item The existing literature utilizes static aggregation for the ensemble block of edRVFL. However, such a static ensemble does not consider the evolving characteristic or the concept drift problem. Recently, a dynamic ensemble algorithm that computes the ensemble weights based on the latest accuracy is proposed for forecasting by \citet{liang2022bayesian}. The output layers of edRVFL can be considered as different models. Hence, the dynamic ensemble can be applied to combine all output layers' forecasts. Therefore, a dynamic ensemble that assigns evolving weights can be combined with edRVFL to boost the performance further.
  \item The tuning process of edRVFL imposes a significant effect on the performance. A layer-wise tuning algorithm is proposed for time series forecasting in \cite{gao2021random}. Such a tuning procedure benefits the diverse and optimal architecture of the edRVFL. However, \citet{gao2021random} only implements a layer-wise tuning algorithm with Bayesian optimization. In the future, more advanced optimization algorithms can be combined with layer-wise tunings, such as evolutionary algorithms \cite{das2010differential}. The marriage of layer-wise tuning and advanced optimization algorithms will develop the RVFLs into auto-ML in the future.
  
  \item Although the random features offer non-linearity and fast computation, the random nature carries redundant information. Therefore, an intelligent selection of the random features owns the strong potential to increase the performance \cite{CHENG2021107826,shi2022weighting}. The inferior features of random layers may deteriorate the performance. The existing works only consider linear feature selection, and pruning techniques \cite{CHENG2021107826,shi2022weighting}. However, there are more advanced feature selection algorithms \cite{chandrashekar2014survey}. The effects of applying different feature selection algorithms on the random features must be studied. 

\item Although RVFL and its variants show superior forecasting, there are still directions worth exploring. For instance, the augmentation of RVFL's random features is not mature yet, although signal decomposition shows its effectiveness. If the decomposition is done correctly, the elements generated are always high-dimensional. In RVFL and deep RVFL, the direct links are connected to the linear output layers. Therefore, effective treatment of such high-dimensional features is critical. Potential solutions can be dimensionality reduction, feature selection and double regularizations. Dimensionality reduction algorithms can be utilized to transform the huge input feature matrix into a low-dimensional space. Then, a linear layer is trained in the low-dimensional space. Feature selection only selects a few best features for the linear layer to learn. As for the double regularizations, different regularizations are imposed on the direct link and random features. If the direct link is of high dimension, its regularization would prefer sparsity.
\item RVFL's forecasting ability on spatial-temporal time series is not investigated yet. The spatial-temporal time series is a temporal sequence of graph signals. However, the conventional version of RVFL is not suitable for graph data. Therefore, it is a promising direction to develop RVFL for graph data. 
\item Recently, multi-label learning has been emerging as an exciting research domain. Therefore, researchers may develop randomized neural networks to handle multi-label data. RVFL model doesn't have enough work to manage the multi-label datasets. The efficient and effective variants of the RVFL model should be developed for multi-label tasks.
\item Unlike supervised learning, unsupervised learning problems doesn't have target variable information. Standard RVFL model needs target information while calculating the final parameters. Unlabeled data are clustered (or grouped) by considering their topological properties or other properties of the data. Therefore, the needful works are required to handle unlabeled data via RVFL model.
\item The community lacks a thorough investigation that compares the performance of randomised neural networks on datasets that are openly accessible, utilising standardised metrics, evaluation procedures, and several datasets. A benchmark for comparing the various randomised architectures is thus needed in the field. This will encourage future efforts to improve a randomised model from various angles, including precision, trustworthiness, and training/inference efficiency.
\item Most of the RVFL-based architectures are based on the offline training wherein all the data is available for the training at once. However, in online scenarios the sequential streaming data needs to be  processed. RVFL models can be adapted to handle such scenarios. Moreover, one can also focus on the development of deep RVFL architectures for the online learning process. As for the edRVFL, the output layers can be trained in an online fashion and the ensemble can be online, too.
\end{itemize}

\section*{Acknowledgement}
This work is supported by the National Supercomputing Mission under DST and Miety,  Govt. of India under Grant No. DST/NSM/R\&D\_HPC \_Appl/2021/03.29, as well as the Department of Science and Technology under Interdisciplinary Cyber Physical Systems (ICPS) Scheme grant no. DST/ICPS/CPS-Individual/2018/276 and Mathematical Research Impact-Centric Support (MATRICS) scheme grant no. MTR/2021/000787. Mr. Ashwani kumar Malik acknowledges the financial support (File no - 09/1022 (0075)/2019-EMR-I) given as scholarship by Council of Scientific and Industrial Research (CSIR),
New Delhi, India. We are  grateful to  IIT Indore for the facilities and support being provided.
\bibliographystyle{elsarticle-num-names}

\bibliography{refs.bib}

\end{document}